%% file: main.tex
\documentclass{article} 
\usepackage{iclr2023_conference,times}
\iclrfinalcopy

\input{utils_general}
\input{utils_math}

\usepackage{hyperref}
\usepackage{url}
\usepackage{multirow}
\usepackage{graphicx}
\usepackage{hyperref}
\usepackage{graphicx}
\usepackage{booktabs}
\usepackage{amssymb} 
\usepackage{pifont}
\usepackage{todo}
\usepackage[normalem]{ulem}
\usepackage{algorithm,algpseudocode}
\usepackage{color}
\usepackage{adjustbox}
\usepackage[most]{tcolorbox}
\usepackage{wrapfig}
\usepackage{cite}
\usepackage{xcolor,colortbl}
\usepackage{xspace}
\algnewcommand{\algorithmicforeach}{\textbf{for each}}
\algdef{SE}[FOR]{ForEach}{EndForEach}[1]
  {\algorithmicforeach\ #1\ \algorithmicdo}
  {\algorithmicend\ \algorithmicforeach}

\newcommand{\irm}{{\textsc{IRMv1}}}
\newcommand{\irmzero}{{\textsc{IRMv0}}}
\newcommand{\sirm}{{\textsc{SparseIRM}}}
\newcommand{\birm}{{\textsc{BIRM}}}
\newcommand{\fishr}{{\textsc{Fishr}}}
\newcommand{\game}{{\textsc{IRM-Game}}}
\newcommand{\rex}{{\textsc{REx}}}
\newcommand{\erm}{{\textsc{ERM}}}
\newcommand{\oracle}{{\textsc{GrayScale}}}
\newcommand{\optim}{{\textsc{Optimum}}}

\newcommand{\irmLSGD}{{\textsc{IRM-LSGD}}}
\newcommand{\irmLALR}{{\textsc{IRM-LALR}}}
\newcommand{\irmSAM}{{\textsc{IRM-SAM}}}

\newcommand{\CMNIST}{{\textsc{Colored-MNIST}}}
\newcommand{\CFMNIST}{{\textsc{Colored-FMNIST}}}
\newcommand{\COBJECT}{{\textsc{Colored-Object}}}
\newcommand{\CM}{{\textsc{Cifar-MNIST}}}
\newcommand{\CELEBA}{{\textsc{CelebA}}}
\newcommand{\IRMGame}{\textsc{IRM-Game}\xspace}
\newcommand{\IRMConsensus}{{\textsc{BLOC-IRM}}}

\newcommand{\IRM}{\text{IRM}\xspace}
\newcommand{\ours}{\IRMConsensus}
\newcommand{\oursTable}{{\textsc{BLOC-IRM}}}

\newcommand{\De}{\mathcal D^{(e)}}
\newcommand{\Le}{\ell^{(e)}}

\newcommand{\Envtr}{\mathcal E_{\mathrm{tr}}}
\newcommand{\RepMap}{\phi_{\boldsymbol \theta}}
\newcommand{\PredMap}{f_{\mathbf w}}

\makeatletter
\newcommand*{\rom}[1]{\expandafter\@slowromancap\romannumeral #1@}
\makeatother

\title{
What Is Missing in IRM Training and Evaluation? Challenges and Solutions
}

\author{Yihua Zhang$^1$, Pranay Sharma$^2$, Parikshit Ram$^3$, Mingyi Hong$^4$, Kush Varshney$^3$, Sijia Liu$^{1,3}$\\
$^1$Michigan State University, $^2$Carnegie Mellon University, $^3$IBM Research, $^4$University of Minnesota
}

\begin{document}

\maketitle

\begin{abstract}
    Invariant risk minimization (IRM) has received increasing attention 
    as a way to acquire environment-agnostic data representations and predictions, and  
    as a principled solution for preventing spurious correlations from being learned and for improving models' out-of-distribution generalization. Yet, recent works have found that the optimality of the originally-proposed IRM optimization ({\irm}) may be compromised in practice or could be  impossible to achieve in some scenarios. Therefore, a series of advanced IRM algorithms have been developed that show practical improvement over {\irm}. In this work, we revisit these recent IRM advancements, and identify and resolve three practical limitations
    in IRM training and evaluation. 
    First, we find that the effect of batch size during training has been chronically overlooked in previous studies, leaving room for further improvement. 
    We propose small-batch training and highlight the improvements over a set of large-batch optimization techniques. 
    Second, we find that improper selection of evaluation environments could give a false sense of invariance for IRM.
    To alleviate this effect, we leverage diversified test-time environments to precisely characterize the invariance of IRM when applied in practice.
    Third, we revisit \citet{ahuja2020invariant}'s proposal to convert IRM into an ensemble game and identify a limitation when a single invariant predictor is desired instead of an ensemble of individual predictors. We propose a new IRM variant to address this limitation 
    based on a novel viewpoint of ensemble IRM games as consensus-constrained bi-level optimization. Lastly, we conduct extensive experiments (covering 7 existing IRM variants and 7 datasets) to justify the practical significance of
    revisiting IRM training and evaluation in a principled manner.
\end{abstract}

\section{Introduction}

Deep neural networks (DNNs) have enjoyed unprecedented success in many real-world applications \citep{he2016deep, krizhevsky2017imagenet, simonyan2014very, sun2014deep}. 
However, experimental evidence \citep{beery2018recognition, de2019causal, degrave2021ai, geirhos2020shortcut, zhang2022nico} suggests that 
DNNs trained with empirical risk minimization (\textbf{ERM}), the most commonly used training method, are prone to reproducing spurious correlations in the training data \citep{beery2018recognition, sagawa2020investigation}. 
This phenomenon causes performance degradation when facing distributional shifts at test time \citep{gulrajani2020search, koh2021wilds, wang2022generalizing, zhou2022domain}. 
In response, the problem of \textit{invariant prediction} arises to enforce the model trainer to learn stable and causal features \citep{beery2018recognition, sagawa2020investigation}.

In pursuit of out-of-distribution generalization, a new model training paradigm, termed invariant risk minimization (\textbf{IRM}) \citep{arjovsky2019invariant}, has received increasing attention to overcome the shortcomings of ERM against distribution shifts.
In contrast to ERM, IRM aims to learn a universal representation extractor, which can elicit an \textit{invariant} predictor across multiple training environments. 
However, different from ERM, the learning objective of IRM is highly non-trivial to optimize in practice. 
Specifically, IRM requires solving a challenging bi-level optimization (\textbf{BLO}) problem with a hierarchical learning structure: invariant representation learning at the \textit{upper-level} and invariant predictive modeling at the \textit{lower-level}. 
Various techniques have been developed to solve IRM effectively, such as \citep{ahuja2020invariant, lin2022bayesian, rame2022fishr, zhou2022sparse} to name a few. 
Despite the proliferation of IRM advancements, several issues in the theory and practice   have also appeared. 
For example,  recent works \citep{rosenfeld2020risks, kamath2021does} revealed the theoretical failure of IRM in some cases.
In particular, there exist scenarios where the optimal invariant predictor is impossible to achieve, and the IRM performance may fall behind even that of ERM. 
Practical studies also demonstrate that the performance of IRM   rely on multiple factors, \textit{e.g.}, model size \citep{lin2022bayesian, zhou2022sparse}, environment difficulty \citep{dranker2021irm, krueger2021out}, and dataset type \citep{gulrajani2020search}.

Therefore, key challenges remain in deploying IRM to real-world applications. 
In this work, we revisit recent IRM advancements and uncover and tackle several pitfalls in IRM training and evaluation, which have so far gone overlooked.
We first identify the large-batch training issue in existing IRM algorithms, which prevents escape from bad local optima during IRM training. Next, we show that evaluation of IRM performance with a \textit{single} test-time environment could lead to an inaccurate assessment of prediction invariance, even if this test environment differs significantly from training environments. 
Based on the above findings, we further develop a novel IRM variant, termed {\textsc{\textbf{BLOC}-IRM}}, by interpreting and advancing the {\game} method \citep{ahuja2020invariant} through the lens of \textbf{BLO} with \textbf{C}onsensus prediction.
Below, we list our \textbf{contributions} (\ding{182}-\ding{185}).

\ding{182} We demonstrate that the prevalent use of large-batch training leaves significant room for performance improvement in IRM, something chronically overlooked in the previous IRM studies with benchmark datasets {\CMNIST} and {\CFMNIST}. 
By reviewing and comparing with 7 state-of-the-art (SOTA) IRM variants  (Table\,\ref{tab: method_overview}),
we show that simply using \textit{small-batch} training improves generalization over a series of  more involved  large-batch optimization enhancements. 

\ding{183} We also show that an inappropriate evaluation metric could give a false sense of invariance to IRM. Thus, we propose an extended evaluation scheme that quantifies both precision and `invariance' across diverse  testing environments.
    
\ding{184} Further, we revisit and advance the {\IRMGame} approach \citep{ahuja2020invariant} through the lens of consensus-constrained BLO. We remove the need for an ensemble (one per training environment) of predictors in {\game} by proposing 
\textbf{\ours} ({\bf BLO} with {\bf C}onsensus \textbf{\IRM}), which produces a single invariant predictor.

\ding{185} 
Lastly, we conduct extensive experiments (on 7 datasets, using diverse model architectures and training environments) to justify the practical significance of our findings and methods. Notably, we conduct experiments on the {\CELEBA} dataset as a new IRM benchmark with realistic spurious correlations. 
We show that {\ours}  outperforms all baselines in nearly all settings.

\subsection{Related Work}

\paragraph{IRM methods.} 
Inspired by the invariance principle
\citep{peters2016causal}, \citet{arjovsky2019invariant} 
define IRM  as a BLO problem, and develop a relaxed single-level formulation, termed {\irm}, for ease of training.
Recently, there has been considerable work to advance IRM techniques. Examples of IRM variants include penalization on the variance of risks or loss gradients across training environments \citep{chang2020invariant, krueger2021out, rame2022fishr, xie2020risk, xu2021learning, xu2022regulatory}, 
domain regret minimization \citep{jin2020domain}, robust optimization over multiple domains  \citep{xu2021learning}, sparsity-promoting invariant learning  \citep{zhou2022sparse},
Bayesian inference-baked IRM \citep{lin2022bayesian},
and  ensemble game over the environment-specific predictors  \citep{ahuja2020invariant}.
We refer readers to Section\,\ref{sec: preliminaries_setup} and Table\,\ref{tab: method_overview} for more details on the IRM methods that we will focus on in this work.

Despite the potential and popularity of IRM, some works have also shown the theoretical and practical limitations of current IRM algorithms.
Specifically, \citet{chen2022pareto, kamath2021does} show that invariance learning via IRM could fail and be worse than ERM in some two-bit environment setups on {\CMNIST}, a synthetic benchmark dataset often used in IRM works. 
The existence of failure cases of IRM is also theoretically shown by  \citet{rosenfeld2020risks} for both linear and non-linear models. Although subsequent IRM algorithms take these failure cases into account, there still exist huge gaps between theoretically desired IRM and its practical variants. For example, \citet{lin2021empirical, lin2022bayesian, zhou2022sparse} found many IRM variants incapable of maintaining graceful generalization on large and deep models.
Moreover, \citet{ahuja2020empirical, dranker2021irm}
demonstrated that  the performance of IRM algorithms could depend on practical details, \textit{e.g.}, dataset size, sample efficiency, and environmental bias strength.
The above IRM limitations   inspire our work to study when and how we can turn the IRM advancements into effective solutions, to gain high-accuracy and stable invariant predictions in practical scenarios.

\paragraph{Domain generalization.}
IRM is also closely related to domain generalization \citep{carlucci2019domain, gulrajani2020search, koh2021wilds, li2019episodic, nam2021reducing, wang2022generalizing, zhou2022domain}. Compared to IRM, 
domain generalization includes a wider range of approaches to improve prediction accuracy against distributional shifts \citep{beery2018recognition, jean2016combining, koh2021wilds}. 
For example, an important line of research is to
improve representation learning by encouraging cross-domain feature resemblance \citep{long2015learning, tzeng2014deep}.
The studies on domain generalization have also been conducted across different learning paradigms, \textit{e.g.}, adversarial learning  \citep{ganin2016domain}, self-supervised learning  \citep{carlucci2019domain}, and meta-learning \citep{balaji2018metareg, dou2019domain}.

\input{sections_SL/preliminaries}

\input{sections_SL/sec_batch_IRM}

\input{sections_SL/sec_multieval_IRM}

\input{sections_SL/sec_BLO_IRM_Game}

\input{sections_SL/sec_experiment}

\vspace*{-1mm}
\section{Conclusion}
In this work, we investigate existing IRM methods and reveal long-standing but chronically overlooked challenges involving IRM training and evaluation, which may lead to sub-optimal solutions and incomplete invariance assessment. As a remedy, we propose small-batch training and multi-environment evaluation. We reexamine the {\IRMGame} method through the lens of consensus-constrained BLO, and develop a novel IRM variant, termed {\ours}. We conducted extensive experiments on 7 datasets and demonstrate that {\ours} consistently improves all baselines.

\section*{Acknowledgement}
The work of Y. Zhang and S. Liu was partially supported by National Science Foundation (NSF) Grant IIS-2207052.
The work of M. Hong was supported by NSF grants CNS-2003033 and CIF-1910385.
The computing resources used in this work were partially supported by the MIT-IBM Watson AI Lab and the Institute for Cyber-Enabled Research (ICER) at Michigan State University.

\section*{Reproducibility Statement}
The authors have made an extensive effort to ensure the reproducibility of algorithms and results presented in the paper. \underline{First}, the details of the experiment settings have been elaborated in Section\,\ref{sec: exp_setup} and Appendix\,\ref{app: exp_setup}. In this paper, seven datasets are studied and the environment generation process for each dataset is described with details in Appendix\,\ref{app: dataset}. The evaluation metrics are also clearly introduced in Section\,\ref{sec: large-batch-improvement}. 
\underline{Second},
eight IRM-oriented methods (including our proposed {\ours}) are studied in this work. 
The implementation details of all the baseline  methods are clearly presented in Appendix\,\ref{app: baseline}, including the hyper-parameters tuning, model  configuration, and   used code bases. For our proposed {\ours}, we include all the implementation details in Section \,\ref{sec: consensus} and Appendix\,\ref{app: bloc_impl}, including training pipeline  in Figure\,\ref{fig: alg_illustration} and the pseudo-code in Algorithm\,\ref{alg: bloc}.
\underline{Third}, all the results are based on 10 independent trials with different random seeds. The standard deviations are also reported to ensure fair comparisons across different methods. 
\underline{Fourth}, codes are available at \url{https:/github.com/OPTML-Group/BLOC-IRM}.

\newpage
\clearpage
{\small
\bibliographystyle{iclr2023_conference}
\bibliography{refs/ref_IRM, refs/ref_SL_adv
}
}

\input{appendix}

\end{document}

%% file: utils_math.tex
\usepackage{amsmath,amsfonts,bm}

\def\eqref#1{(\ref{#1})}

\def\1{\bm{1}}

\DeclareMathAlphabet{\mathsfit}{\encodingdefault}{\sfdefault}{m}{sl}
\SetMathAlphabet{\mathsfit}{bold}{\encodingdefault}{\sfdefault}{bx}{n}

\DeclareMathOperator*{\argmin}{arg\,min}

\DeclareMathOperator*{\minimize}{\text{minimize}}

\DeclareMathOperator*{\st}{\text{subject to}}
\DeclareMathAlphabet\mathbfcal{OMS}{cmsy}{b}{n}
\newcommand{\Def}[0]{\mathrel{\mathop:}=}

\newcommand{\bx}{\mathbf{x}}

\newcommand{\bw}{\mathbf{w}}

\newcommand{\btheta}{\boldsymbol{\theta}}

%% file: sections_SL/preliminaries.tex
\section{Preliminaries and Setup}
\label{sec: preliminaries_setup}
In this section, we  introduce the basics of IRM and provide an overview of our IRM case study.

\noindent \textbf{IRM formulation.}
In the original IRM framework     
\citet{arjovsky2019invariant}, 
consider a supervised learning paradigm, with  datasets $\{ \De \}_{e \in {\Envtr}}$ collected from $N$  training environments $\Envtr = \{1,2,\ldots, N \}$. 
The training samples in   $\De$ (corresponding to the environment $e$) are of the form $(\bx,y) \in \mathcal X \times \mathcal Y$, where $\mathcal X$ and $\mathcal Y$ are, respectively, the raw feature space and the label space.
IRM aims to find an \textit{environment-agnostic data representation}
$\RepMap: \mathcal X \to \mathcal Z$, which elicits an \textit{invariant prediction} $\PredMap: \mathcal Z \to \mathcal Y$ that is simultaneously optimal for all environments. 
Here $\boldsymbol \theta$ and $\mathbf w$ denote model parameters to be learned, and $\mathcal Z$ denotes the representation space.
Thus, IRM yields an \textit{invariant predictor} $\PredMap \circ \RepMap: \mathcal X \to \mathcal Y$ that can generalize to unseen test-time environments $\{ \De \}_{e \notin {\Envtr}}$. 
Here $\circ$ denotes function composition, \textit{i.e.}, $\PredMap \circ \RepMap (\cdot) = \PredMap ( \RepMap (\cdot))$.
We will use $\mathbf w \circ \boldsymbol \theta$ as a shorthand for $\PredMap \circ \RepMap$.
IRM constitutes the following BLO problem:

\vspace*{-4.5mm}
{\small 
\begin{align}
    \begin{array}{ll}
\displaystyle \minimize_{\boldsymbol \theta}         & { \sum_{e \in \Envtr } \Le ( \mathbf w^*(\boldsymbol \theta) \circ \boldsymbol \theta)};  \quad 
  \st       ~~ \mathbf w^*(\boldsymbol \theta) \in \displaystyle \argmin_{\bar {\mathbf w}} \Le (\bar {\mathbf w} \circ \boldsymbol \theta),~~\forall e \in {\Envtr},
    \end{array}
    \label{eq: prob_IRM_ori}
    \tag{IRM}
\end{align}}%
where $\Le (\mathbf w \circ \boldsymbol \theta)$ is the per-environment training loss of the predictor $\mathbf w \circ \boldsymbol \theta$ under  $\De$. 
Clearly, \ref{eq: prob_IRM_ori} involves two optimization levels that are coupled  through the lower-level solution $\mathbf w^*(\boldsymbol \theta) $. 
Achieving the desired invariant prediction requires 
the solution sets of the individual lower-level problems $\{ \argmin_{\bar {\mathbf w}} \Le (\bar {\mathbf w} \circ \boldsymbol \theta), e \in \mathcal{E}_{tr} \}$ to be non-singleton. 
However, BLO problems with non-singleton lower-level solution sets are significantly more challenging \citep{liu2021investigating}.
To circumvent this difficulty, 
\citet{arjovsky2019invariant} relax  \eqref{eq: prob_IRM_ori} into a   single-level optimization problem (\textit{a.k.a.}, IRMv1):

\vspace*{-4.5mm}
{\small \begin{align}
    \begin{array}{ll}
\displaystyle \minimize_{\boldsymbol \theta}         & \sum_{e \in \Envtr } [ \Le ( \boldsymbol \theta) + \gamma \| \nabla_{w   |   w = 1.0} \Le ( w \circ \boldsymbol \theta) \|_2^2 ],
    \end{array}
    \label{eq: prob_IRMv1_ori}
    \tag{IRMv1}
\end{align}}%
where $\gamma > 0$ is a regularization parameter and $\nabla_{w|w = 1.0} \Le$ denotes the gradient 
of $\Le$ with respect to $w$, computed at $w = 1.0$. 
Compared with \ref{eq: prob_IRM_ori}, \ref{eq: prob_IRMv1_ori} 
is restricted to {\em linear} invariant predictors,
and penalizes 
the deviation of individual environment losses from stationarity
to approach the lower-level optimality in \eqref{eq: prob_IRM_ori}. 
\ref{eq: prob_IRMv1_ori} uses the fact that a scalar predictor ($w = 1.0$) is equivalent to a linear predictor.
Despite the practical simplicity of \eqref{eq: prob_IRMv1_ori}, it may fail to achieve the desired invariance \citep{chen2022pareto, kamath2021does}. 

\begin{figure}[!t]
\begin{minipage}[th]{\textwidth}
\begin{minipage}[t]{0.55\textwidth}
\makeatletter\def\@captype{table}
\centering
\caption{\footnotesize{Summary of the $7$ existing IRM variants considered in this work, and the proposed {\ours} method (see Section\,\ref{sec: consensus}). 
We also list the $7$ benchmark datasets used to evaluate IRM performance, namely,
{\CMNIST} (CoM), {\CFMNIST} (CoF), {\CM} (CiM), {\COBJECT} (CoO), \textsc{CelebA} (CA), \textsc{PACS} (P) and \textsc{VLCS} (A). 
The symbols `\ding{51}' signifies the dataset used in the specific reference.
\vspace*{1mm}
}}
\label{tab: method_overview}
\resizebox{\textwidth}{!}{%
\begin{tabular}{c|c|ccccccc|c}
\toprule[1pt]
\midrule
\multirow{2}{*}{\begin{tabular}[c]{@{}c@{}}IRM\\ Method\end{tabular}} & \multirow{2}{*}{Venue} & \multicolumn{7}{c|}{Datasets} & \multirow{2}{*}{Reference} \\ 
 &  & CoM & CoF & CiM & CoO & CA & P & V &  \\
 \midrule
{\irm} & arXiv & \ding{51} &  
 
& 
&  
&  
&  
&  
& \citep{arjovsky2019invariant} \\
{\irmzero} & N/A & 
 
& 
 
& 
 
& 

& 
 
& 

& 
 
& This Work \\
{\game} & ICML & \ding{51} & \ding{51} &  &  &  &  &  & \citep{ahuja2020invariant} \\
{\rex} & ICML & \ding{51} & \ding{51} &  &  &  & \ding{51} & \ding{51} & \citep{krueger2021out} \\
{\birm} & CVPR & \ding{51} &  & \ding{51} & \ding{51} &  &  &  & \citep{lin2022bayesian} \\
{\sirm} & ICML & \ding{51} &  & \ding{51} & \ding{51} &  &  &  & \citep{zhou2022sparse} \\
{\fishr} & ICML & \ding{51} &  &  &  &  & \ding{51} & \ding{51} & \citep{rame2022fishr} \\
\midrule
{Ours} & N/A & \ding{51} & \ding{51} & \ding{51} & \ding{51} & \ding{51} & \ding{51} & \ding{51} & This Work\\
\midrule
\bottomrule[1pt]
\end{tabular}%
}
\end{minipage}
\hfill
\begin{minipage}[t]{0.4\textwidth}
\makeatletter\def\@captype{table}
\centering
\caption{\footnotesize{Dataset setups. `Invariant' and `Spurious' represent the core   and      spurious features. `Env1' and `Env2' are environments  with different spurious correlations.
}}
\label{tab: dataset}
\resizebox{\textwidth}{!}{%
\begin{tabular}{c|cccc}
\toprule[1pt]
\midrule
Dataset        & Invariant  & Spurious   & Env 1 & Env 2 
\\
\midrule
CoM  & Digit & Color      
& \includegraphics[width=2em, height=2em]{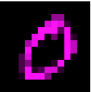} \includegraphics[width=2em, height=2em]{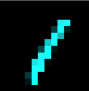}       
& \includegraphics[width=2em, height=2em]{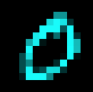} \includegraphics[width=2em, height=2em]{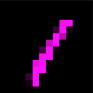}     
\\
CoF & Object     & Color      
& 
\includegraphics[width=2em, height=2em]{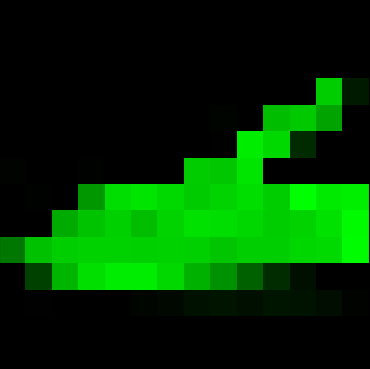} \includegraphics[width=2em, height=2em]{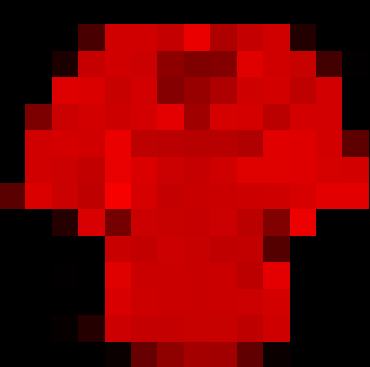}
&
\includegraphics[width=2em, height=2em]{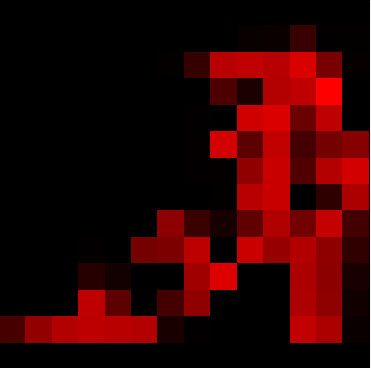} \includegraphics[width=2em, height=2em]{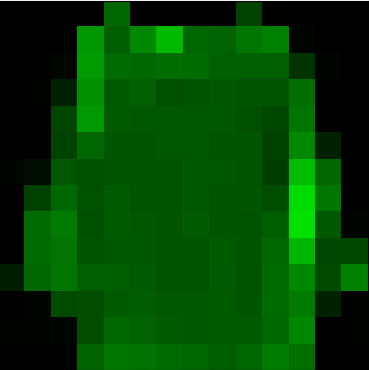}
\\
CiM    & CIFAR      & MNIST      
& \includegraphics[width=2em, height=2em]{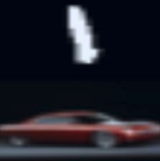} \includegraphics[width=2em, height=2em]{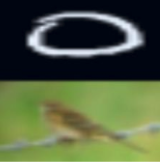}
& \includegraphics[width=2em, height=2em]{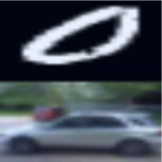} \includegraphics[width=2em, height=2em]{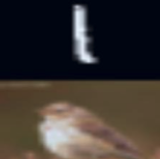} 
\\
CoO & Object     & Color      
& \includegraphics[width=2em, height=2em]{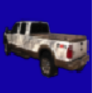} \includegraphics[width=2em, height=2em]{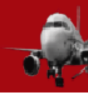}         
& \includegraphics[width=2em, height=2em]{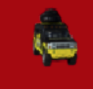} \includegraphics[width=2em, height=2em]{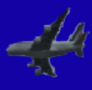}
\\
CA & Smiling & Hair Color 
& \includegraphics[width=2em, height=2em]{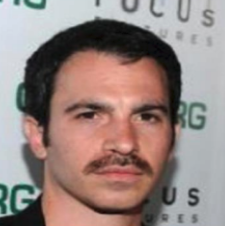} \includegraphics[width=2em, height=2em]{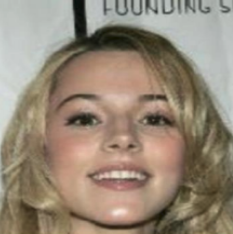}          
& \includegraphics[width=2em, height=2em]{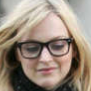} \includegraphics[width=2em, height=2em]{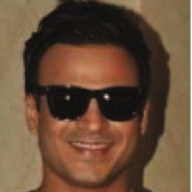}
\\
P & Object & Texture 
& \includegraphics[width=2em, height=2em]{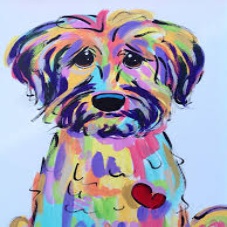} \includegraphics[width=2em, height=2em]{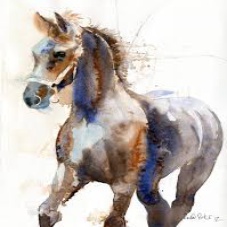}     
& \includegraphics[width=2em, height=2em]{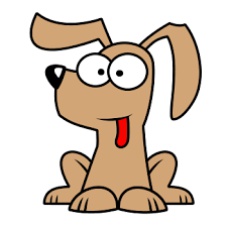} \includegraphics[width=2em, height=2em]{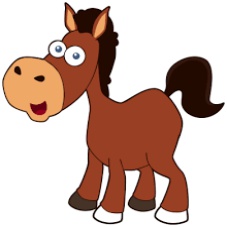}           
\\
V & Object & Environment 
& \includegraphics[width=2em, height=2em]{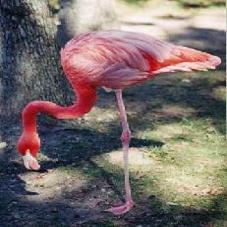} \includegraphics[width=2em, height=2em]{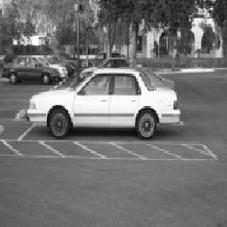} 
& \includegraphics[width=2em, height=2em]{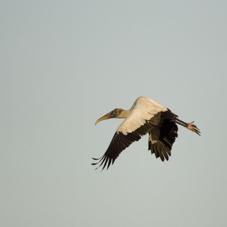} \includegraphics[width=2em, height=2em]{figures/dataset_demo/VLCS-11.jpg} \\
\midrule
\bottomrule[1pt]
\end{tabular}%
}
\end{minipage}
\end{minipage}
\vspace*{-4mm}
\end{figure}

\noindent \textbf{Case study of IRM methods.}
As illustrated above, the  objective of \ref{eq: prob_IRM_ori} is difficult to optimize, while \ref{eq: prob_IRMv1_ori} only provides a sub-optimal solution. Subsequent advances have attempted to reduce this gap. In this work, we focus on 7 popular IRM variants and evaluate their invariant prediction performance over 7 datasets. \textbf{Table\,\ref{tab: method_overview}} and \textbf{Table\,\ref{tab: dataset}} respectively summarize the IRM methods   and the datasets considered in this work. 
We survey
the most representative and effective IRM variants in the literature, which will also serve as our baselines in performance comparison. 

Following Table\,\ref{tab: method_overview},
we first introduce the 
\textbf{{\irmzero}} variant, a generalization of   \ref{eq: prob_IRMv1_ori}, by relaxing its assumption of linearity of the predictor $\bw$, yielding

\vspace*{-6mm}
{\small 
\begin{align}
    \begin{array}{ll}
\displaystyle \minimize_{\bw, \boldsymbol \theta}         & \sum_{e \in \Envtr } [ \Le ( \bw \circ \boldsymbol \theta) + \gamma \| \nabla_{\bw} \Le ( \bw \circ \boldsymbol \theta) \|_2^2 ].
    \end{array}
    \label{eq: prob_IRMv0_ori}
    \tag{IRMv0}
\end{align}}%
Next, we consider the risk extrapolation method \textbf{\rex} \citep{krueger2021out}, an important baseline based on distributionally robust optimization for group shifts \citep{sagawa2019distributionally}. 
Furthermore, inspired by the empirical findings  that the  performance of IRM  could  be sensitive to model size \citep{choe2020empirical, gulrajani2020search}, 
we choose the SOTA methods Bayesian IRM (\textbf{\birm}) \citep{lin2022bayesian} and sparse IRM  (\textbf{\sirm}) \citep{zhou2022sparse}, both of which show improved 
performance with large models.  
Also, we consider the  SOTA method \textbf{\fishr}
~\citep{rame2022fishr}, which modifies IRM to  penalize   the domain-level gradient variance in single-level risk minimization. \textbf{\fishr} provably matches both
domain-level risks and Hessians. 
Lastly, we include \textbf{\IRMGame} \citep{ahuja2020invariant} as a special variant of  IRM. Different from the other methods which seek an invariant predictor, {\IRMGame} endows each environment with a predictor, and leverages this \textit{ensemble} of predictors to achieve invariant representation learning. This is in contrast to other existing works which seek an invariant predictor. Yet, we show in Section\,\ref{sec: consensus} that {\IRMGame} can be interpreted through the lens of consensus-constrained BLO and generalized for invariant prediction. 
We also highlight that diverse
datasets are considered in this work (see Table\,\ref{tab: dataset}) to benchmark IRM's performance. 
More details on dataset selections can be found in Appendix\,\ref{app: dataset}. 

%% file: sections_SL/sec_batch_IRM.tex
\section{Large-batch Training Challenge and Improvement}
\label{sec: large-batch-improvement}

In this section, we demonstrate and resolve the  large-batch training challenge  in  current IRM implementations (Table\,\ref{tab: method_overview}).

\begin{wrapfigure}{r}{0.35\textwidth}
\centering
\vspace*{-5mm}
\resizebox{.32\textwidth}{!}{
\centering
\includegraphics[width=.32\textwidth]{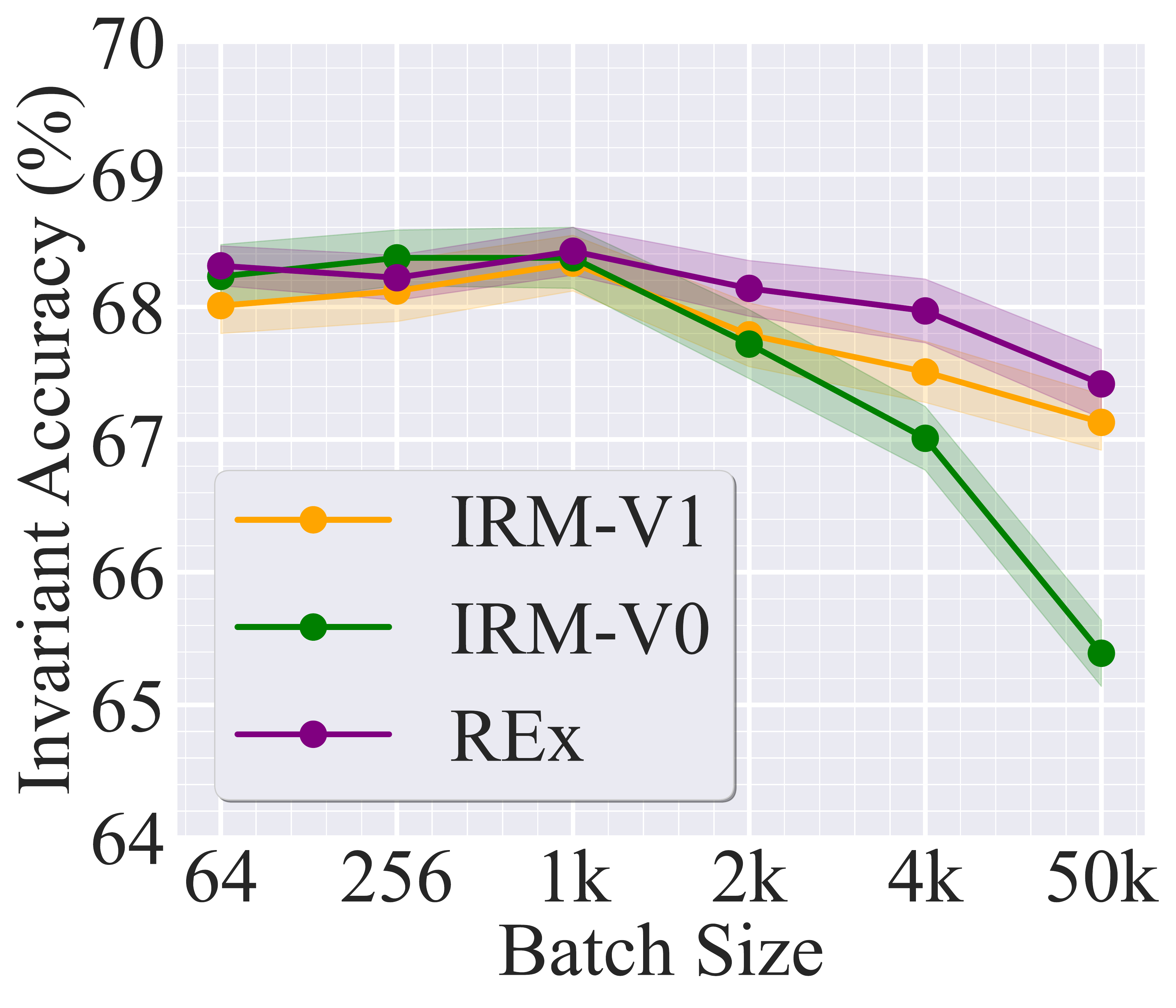}
}
\vspace*{-5mm}
\caption{\footnotesize{The performance of three IRM methods ({\irm}, {\irmzero}, and {\rex}) vs. batch-size under {\CMNIST}. The full   batch-size is 50k.
}}
\vspace*{-1.5em}
\label{fig: acc_bs}
\end{wrapfigure}
\noindent\textbf{Large-batch optimization causes  instabilities of IRM training.}
Using very large-size batches for model training can result in the model getting trapped near a bad local optima \citep{keskar2016large}. This happens as a result of the lack of stochasticity in the training process, and is known to exist even in the ERM paradigm \citep{goyal2017accurate, you2017large}. Yet, nearly all the existing IRM  methods follow the training setup of {\irm} \citep{arjovsky2019invariant}, which used the \textit{full-batch} gradient descent (GD) method rather than the \textit{mini-batch} stochastic gradient descent (SGD) for IRM training over {\CMNIST} and {\CFMNIST}. 
In the following, we show that large-batch training 
might give a false impression of the relative ranking of IRM   performances.

We start with an exploration 
of the impact of batch size on the invariant prediction accuracy  of existing IRM methods under {\CMNIST}. Here the invariant prediction accuracy refers to the averaged accuracy of the invariant predictor applied to diverse test-time environments. We defer its formal description   to Section\,\ref{sec: evaluation}. 
\textbf{Figure \,\ref{fig: acc_bs}} shows the invariant prediction accuracy of  three IRM methods 
{\irm}, {\irmzero}, and {\rex} vs. the data batch size (see Figure\,\ref{fig: bs_full} for results of other IRM variants and Figure\,\ref{fig: acc_bs_cfmnist} for {\CFMNIST}). Recall that the full batch size (50k) was used in  the existing IRM implementations  \citep{arjovsky2019invariant, krueger2021out}. 
As we can see, in the full-batch setup, IRM methods lead to widely different  invariant prediction accuracies, where {\rex} and {\irm} significantly outperform {\irmzero}.
In contrast, in the small-batch case (with size 1k), the discrepancy in accuracy across methods vanishes. 
We see that \ref{eq: prob_IRMv0_ori} can be as effective as \ref{eq: prob_IRMv1_ori} and  other IRM variants (such as {\rex}) \textit{only if} an appropriate \textit{small} batch size is used. 

Empirical evidence in Figure\,\ref{fig: acc_bs} shows that large-batch IRM training  is less effective than small-batch. This is aligned with the observations in ERM \citep{you2017scaling, you2018imagenet, you2019large}, where the lack of stochasticity makes the optimizer difficult to escape from a sharp local minimum. We also justify this issue by visualizing the loss landscapes in Figure\,\ref{fig: trajactory_bs}. Notably, the small-batch training enables {\irm} to  converge to 
a local optimum with a flat loss landscape, indicating better generalization \citep{keskar2016large}. 

\noindent\textbf{Small-batch training is effective versus a zoo of large-batch optimization enhancements.}
To mitigate the large-batch IRM training issue,  we next investigate  the effectiveness of both small-batch training and a zoo of large-batch optimization enhancements.
Inspired by large-batch training techniques to scale up ERM, we consider Large-batch SGD (\textbf{LSGD}) \citep{goyal2017accurate} and {L}ayer-wise Adaptive Learning Rate  (\textbf{LALR})  \citep{you2017scaling, you2018imagenet, you2019large, zhang2022distributed}. 
Both  methods aim to smoothen the optimization trajectory by improving either the learning rate scheduler or the quality of initialization. Furthermore, we adopt sharpness-aware minimization (\textbf{SAM}) \citep{foret2020sharpness} as another possible large-batch training solution to explicitly penalize the sharpness of the loss landscape. We integrate the above optimization techniques with IRM, leading to the variants \textbf{\irmLSGD}, \textbf{{\irmLALR}}, and \textbf{{\irmSAM}}. See Appendix\,\ref{app: large_impl} for more details.

\begin{wraptable}{r}{.41\textwidth}
\centering
\vspace*{-2em}
\caption{\footnotesize{Prediction accuracy of IRM methods on {\CMNIST} 
using the original large-batch implementation (`Original'), the large-batch optimization-integrated implementations (`LSGD/LALR/SAM'), and the small-batch training recipe (`Small').
}}
\label{tab: motiv_bs}
\resizebox{.41\textwidth}{!}{%
\begin{tabular}{c|ccccc}
\toprule[1pt]
\midrule
Method & \textcolor{black}{Original} & \textcolor{black}{LSGD} & \textcolor{black}{LALR} & \textcolor{black}{SAM} & \textcolor{black}{\textbf{Small}} \\ \midrule
\irm        & 67.13 & 67.31 & 67.44 & 67.79 & \textbf{68.33} \\
\irmzero    & 65.39 & 66.42 & 66.76 & 66.99 & \textbf{68.37} \\
\game       & 65.69 & 65.82 & 65.47 & 66.23 & \textbf{67.73} \\
\rex        & 67.42 & 67.53 & 67.59 & 67.82 & \textbf{68.42} \\
\birm       & 67.93 & 67.99 & 68.21 & 68.32 & \textbf{68.71} \\
\sirm       & 67.72 & 67.85 & 67.99 & 68.13 & \textbf{68.81} \\
\fishr      & 67.88 & 67.82 & 67.93 & 68.11 & \textbf{68.69} \\
\midrule
Average     & 67.02 & 67.25 & 67.34 & 67.63 & \textbf{68.44} \\
\midrule
\bottomrule[1pt]
\end{tabular}%
}
\vspace*{-1em}
\end{wraptable}
In Table\,\ref{tab: motiv_bs}, we compare the performance of  the simplest small-batch IRM training with that of those large-batch optimization technique-integrated IRM variants (\textit{i.e.}, `LSGD/LALR/SAM' in the Table).
As we can see,  the use of large-batch optimization techniques indeed improves the prediction accuracy over the original IRM implementation. We also observe that the use of SAM for IRM is consistently better than LALR and LSGD, indicating the promise of SAM to scale up IRM with a large batch size. Yet,  the small-batch training  protocol consistently outperforms large-batch training
across all the IRM variants (see the column `Small'). 
Additional experiment results in Section\,\ref{sec: exp} show that small-batch IRM training is effective across datasets, and promotes the invariance achieved by all methods.

%% file: sections_SL/sec_multieval_IRM.tex
\section{Multi-environment Invariance Evaluation}
\label{sec: evaluation}

In this section, we revisit the evaluation metric used in   existing IRM methods, and
show that expanding the diversity of test-time environments  would  improve the accuracy of invariance assessment.

Nearly all the existing IRM methods (including those listed in Table\,\ref{tab: method_overview}) follow  the evaluation pipeline used in the vanilla IRM framework \citep{arjovsky2019invariant}, which assesses the performance of the learned invariant predictor on a single unseen test environment. This test-time environment is significantly different from
train-time environments.
For example, {\CMNIST} \citep{arjovsky2019invariant} suggests a principled way to define two-bit environments,
widely-used for  IRM dataset curation. 
Specifically, the {\CMNIST} task is to predict the label of the handwritten digit groups (digits 0-4 for group 1 and digits 5-9 for group 2). The digit number is also spuriously correlated with the digit color (Table\,\ref{tab: dataset}). This spurious correlation is controlled by an \textit{environment bias parameter} $\beta$, which specifies different data environments with different levels of spurious correlation\footnote{In the two-bit environment, there exists another environment parameter $\alpha$ that controls the label noise level. 
}. 
In \citep{arjovsky2019invariant}, $\beta = 0.1$ and $\beta = 0.2$ are used to define  \textit{two training environments}, which sample the color ID by flipping the digit group label with probability $10\%$ and $20\%$, respectively.
At \textit{test time}, 
the invariant accuracy is   evaluated on a single, unseen environment with $\beta = 0.9$.
 
However, the prediction accuracy of IRM could be sensitive to the choice of test-time environment (\textit{i.e.}, the value of $\beta$).
For the default test environment $\beta = 0.9$, the predictor performance of three representative IRM methods ({\irm}, {\game}, {\fishr}) ranked from high to low is {\game}$>${\fishr}$>${\irm}. Given this apparent ranking, we explore more diverse test-time environments, generated by $\beta \in \Omega \Def \{0.05, 0.1, \ldots, 0.95 \}$. 
 \begin{wrapfigure}{r}{.56\textwidth}
    \centering
    \begin{tabular}{cc}
    \hspace*{-1em}
    \includegraphics[height=.24\textwidth]{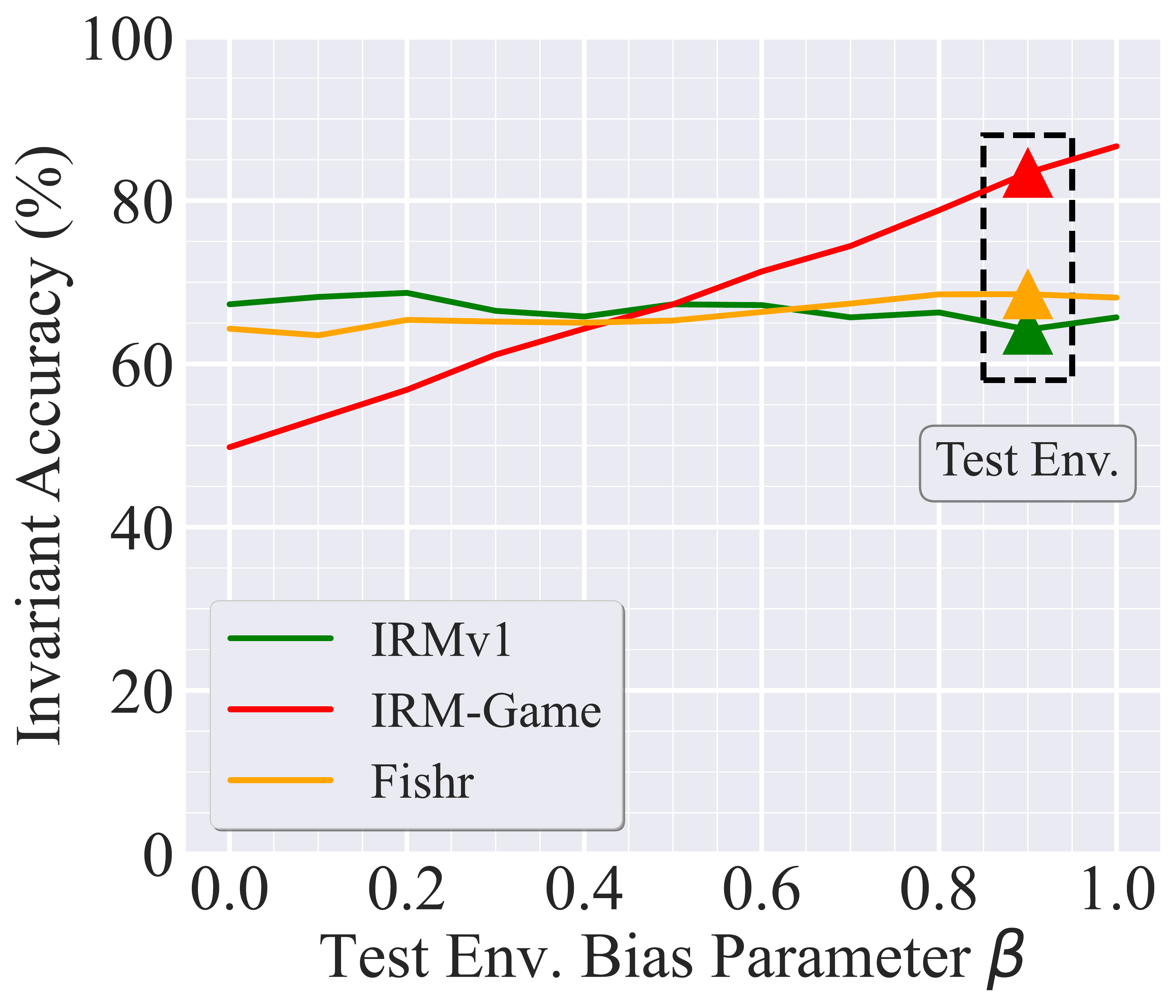}
    & \hspace*{-1.6em} \includegraphics[height=.24\textwidth]{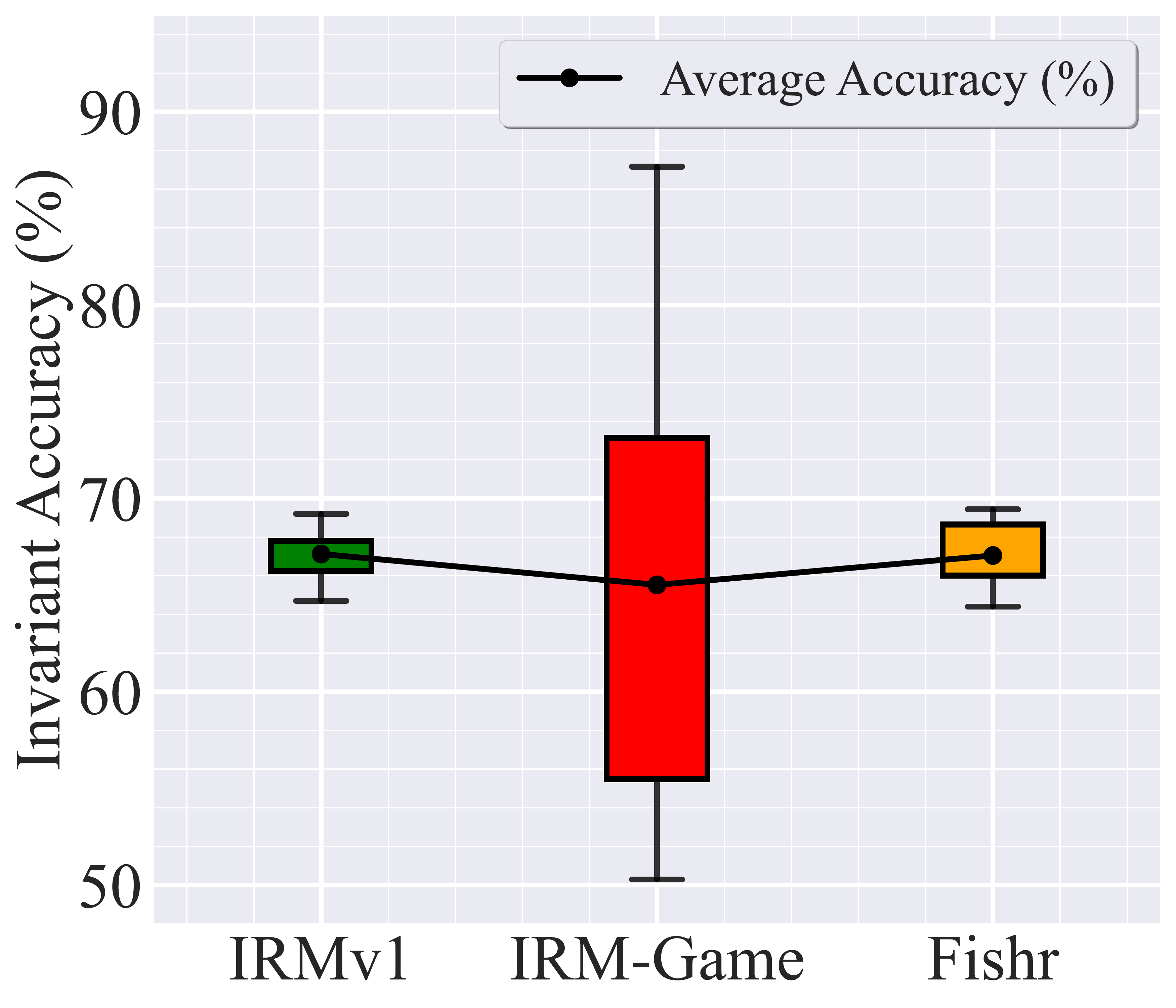} \vspace*{-.5em} \\
    \footnotesize{(A)} & \hspace*{1em} \footnotesize{(B)} 
    \end{tabular}
    \vspace*{-5mm}
    \caption{\footnotesize{Performance comparison  of IRM variants {{\irm}}, {\game}, and {\fishr} on {\CMNIST}. (A) Evaluation in different test-time environments (corresponding to different $\beta$). $\beta$ values used by the two training environments are $0.1, 0.2$ respectively. 
    The conventional evaluation is done with the test environment $\beta = 0.9$ (see `$\blacktriangle$').
    (B) Box plots of prediction accuracies over  diverse test environments corresponding to $\beta \in \{ 0.05, 0.1, \ldots, 0.95 \}$. {\irm} achieves the best average accuracy (67.13\%), followed by {\fishr} (67.05\%) and {\game} (65.53\%).
    }}
    \vspace*{-1.5em}
    \label{fig: motiv_eval}
\end{wrapfigure}
Although the train-time bias parameters $ \{ 0.1, 0.2\}$ belong to $\Omega$, test data is generated afresh, different from training data. We see in Figure\,\ref{fig: motiv_eval}A that the superiority of {\game}  at $\beta = 0.9$ vanishes for smaller $\beta$. Consequently, for invariant prediction evaluated in other testing environments (e.g., $\beta < 0.4$), the performance ranking of the same methods becomes {\irm}$>${\fishr}$>${\game}. This mismatch of results suggests we measure the `invariance' of IRM methods against diverse test environments. Otherwise, evaluation with single $\beta$ could give a false sense of invariance.
In Figure\,\ref{fig: motiv_eval}B, we present the box plots of prediction accuracies for IRM variants, over the diverse set of testing environments $( \beta \in \Omega )$. Evidently, {\irm}, the oldest (sub-optimal) IRM method, yields the least variance of invariant prediction accuracies and the best average prediction accuracy, compared to both {\game} and {\fishr}.
To summarize, the new evaluation method, with diverse test environments, enables us to make a fair comparison of IRM methods implemented in different training environment settings.
Unless specified otherwise, we use the multi-environment evaluation method throughout this work.

%% file: sections_SL/sec_BLO_IRM_Game.tex
\section{Advancing {\game} via Consensus-constrained BLO}
\label{sec: consensus}

In this section, we revist and advance a special IRM variant, {\game} \citep{ahuja2020invariant}, which endows each individual environment with a separate prediction head and converts IRM  into an ensemble game over these multiple predictors.

\noindent\textbf{Revisiting {\game}.}
We first introduce the setup of {\game} following   notations used  in Section\,\ref{sec: preliminaries_setup}. The most essential difference between {\game} and the vanilla \ref{eq: prob_IRM_ori} framework is that the former assigns each environment with an individual classifier $\mathbf w^{(e)}$, and then relies on the ensemble of these individual predictors, \textit{i.e.},  $\frac{1}{N}  \sum_{e \in \Envtr } (  \mathbf w^{(e)}  \circ \boldsymbol \theta )$, for inference.  
{\game} is in a sharp contrast to \ref{eq: prob_IRM_ori}, where an environment-agnostic prediction head $\mathbf w^*$ 
simultaneously optimizes the losses across all environments. Therefore, we raise the following question:  \textit{Can {\game} learn an invariant predictor?
}

Inspired by the above question, 
we explicitly enforce invariance by imposing a \textit{consensus prediction constraint}  $\mathcal C \Def \{ \left( \bar {\mathbf w}^{(1)}, \bar {\mathbf w}^{(2)}, \ldots \bar {\mathbf w}^{(N)} \right) \mid \bar {\mathbf w}^{(1)} =  \ldots =\bar {\mathbf w}^{(N)} \}$ and integrate it with {\game}. 

Here, 
$\bar {\mathbf w}^{(e)}$   denotes the prediction head for the $e$-th  environment. Based on the newly-introduced constraint, the ensemble prediction head $ \frac{1}{N}  \sum_{e \in \Envtr }   \mathbf w^{(e)}  $  can be   interpreted as the \textit{average consensus} over     $N$ environments:
$
\mathbf w^* \Def \frac{1}{N}  \sum_{e \in \Envtr }   \mathbf w^{(e)} = \argmin_{  \{  \bar {\mathbf w}^{(e)} \}_e \in \mathcal C  }  \sum_{e \in \Envtr } \|\bar {\mathbf w}^{(e)} - \mathbf w^{(e)} \|_2^2
$.
With the above consensus interpretation, we can then  cast the  invariant predictor-baked {\game} as a consensus-constrained BLO problem, extended from \eqref{eq: prob_IRM_ori}:

\vspace*{-4.5mm}
{\small 
\begin{align}
    \begin{array}{ll}
\displaystyle \minimize_{\boldsymbol \theta}         & \sum_{e \in \Envtr } \Le (  \mathbf w^*(\btheta) \circ \boldsymbol \theta) \\
  \st        &   
\text{\textbf{(I)}: }\mathbf w^{(e)}(\boldsymbol \theta) \in \displaystyle \argmin_{
   \bar {\mathbf w}^{(e)}  }  \Le (\bar{\mathbf w}^{(e)} \circ \boldsymbol \theta), ~ \forall e \in \Envtr, \vspace*{0.5mm}\\
   & \text{\textbf{(II)}: } \mathbf w^*(\btheta) = \frac{1}{N}  \sum_{e \in \Envtr }   \mathbf w^{(e)} (\btheta).
    \end{array}
    \label{eq: prob_IRM_game_invariance}
\end{align}}%
The above
contains two lower-level problems: \textbf{(I)} per-environment risk minimization, and \textbf{(II)}  projection  onto the consensus  constraint ($\{ {\mathbf w}^{(e)} \} \in \mathcal C$).
The incorporation of (II) is intended to ensure the use of invariant prediction head $\mathbf w^*(\btheta)$ in the upper-level optimization problem of \eqref{eq: prob_IRM_game_invariance}. 

\noindent\textbf{Limitation of \eqref{eq: prob_IRM_game_invariance} and  {\IRMConsensus}.}
In \eqref{eq: prob_IRM_game_invariance}, the  introduced \textit{consensus-constrained} lower-level  problem  might compromise the optimality of the lower-level solution $\mathbf w^*(\btheta)$ to the per-environment {(unconstrained)} risk minimization problem \textbf{(I)}, \textit{i.e.},
violating the per-environment \textit{stationarity} $\| \nabla_{{\mathbf w}} \Le (\mathbf w^*(\btheta) \circ \boldsymbol \theta) \|_2^2$.
Figure\,\ref{fig: stationary_loss} justifies this side effect.
As we can see, the per-environment stationarity is hardly attained at the consensus prediction when solving \eqref{eq: prob_IRM_game_invariance}.
This is not surprising since a constrained optimization solution might not be a stationary solution to minimizing the (unconstrained) objective function.
To alleviate this limitation,  we improve \eqref{eq: prob_IRM_game_invariance} by explicitly promoting the per-environment stationarity  $\| \nabla_{\bw} \Le ( \mathbf w^*(\btheta) \circ \boldsymbol \theta) \|_2^2$ in its upper-level problem through optimization over $\btheta$. This leads to \textbf{\ours} ({\bf BLO} with {\bf C}onsensus \textbf{\IRM}):

\vspace*{-5mm}
{\small 
\begin{align}
    \begin{array}{ll}
\displaystyle \minimize_{\boldsymbol \theta}         & \sum_{e \in \Envtr }  \left [ \Le ( \mathbf w^{*}(\boldsymbol \theta) \circ \boldsymbol \theta) + \gamma  \| \nabla_{\mathbf w} \Le (\mathbf w^*(\btheta) \circ \boldsymbol \theta) 
\|_2^2 \right ] \\
  \st        &  \text{Lower-level problems \textbf{(I)} and \textbf{(II)} in \eqref{eq: prob_IRM_game_invariance}},
    \end{array}
    \label{eq: prob_IRM_game_consensus}
    \tag{{\IRMConsensus}}
\end{align}}%
where $\gamma > 0$ is a regularization parameter like \ref{eq: prob_IRMv0_ori}.
Assisted by the (upper-level) prediction stationarity regularization,  the consensus prediction \textbf{(II)} indeed  simultaneously minimizes  the risks of all the environments, supported by the empirical evidence that the convergence of $\| \nabla_{\bw} \Le ( \mathbf w^*(\btheta) \circ \boldsymbol \theta) \|_2^2$ towards $0$ along each environment's optimization path (see Figure\,\ref{fig: stationary_loss}).

Further, we elaborate on how the \ref{eq: prob_IRM_game_consensus}  problem 
can be effectively solved using an ordinary BLO solver.
First, it is worth noting that although both \eqref{eq: prob_IRM_ori} and  {\IRMConsensus} are BLO problems, the latter is   easier to solve since the lower-level constraint \textbf{(I)} is unconstrained and separable over environments,  and the consensus operation \textbf{(II)} is linear. 
Based on these characteristics, the  implicit gradient   $\frac{d \mathbf w^*(\btheta)}{ d \btheta}$ can be directly computed as

\vspace*{-4.5mm}
{\small 
\begin{align}
    \frac{d \mathbf w^*(\btheta)}{ d \btheta} = \frac{1}{N}  \sum_{e \in \Envtr }  \frac{d  \mathbf w^{(e)} (\btheta)}{d \btheta},  ~~\text{subject to}~~ \mathbf w^{(e)}(\boldsymbol \theta) \in \displaystyle \argmin_{
   \bar {\mathbf w}^{(e)}  }  \Le (\bar{\mathbf w}^{(e)} \circ \boldsymbol \theta).
   \label{eq: prob_IG_uncons}
\end{align}}%

\begin{wrapfigure}{r}{.51\textwidth}
\vspace*{-5mm}
\centering
\includegraphics[width=.51\textwidth]{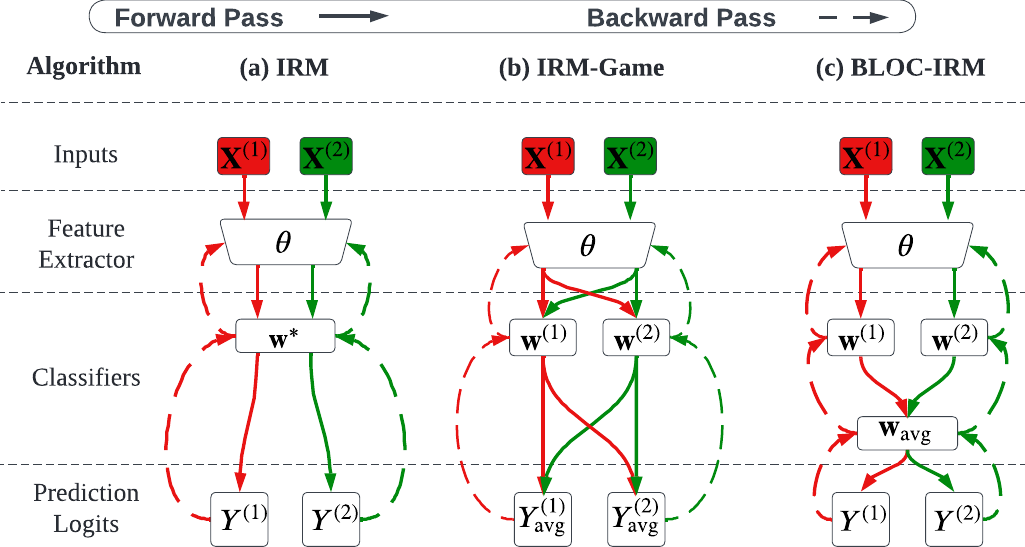}
\vspace*{-1.5em}
\caption{\footnotesize{Schematic overview of {\ours} over two training environments (red and green), and its   comparison to  \textsc{IRM} and {\game}.
}}
\vspace*{-2em}
\label{fig: alg_illustration}
\end{wrapfigure}
Since the above lower-level problem is \textit{unconstrained}, we can call the standard $\argmin$ differentiating method, such as implicit function approach \citep{gould2016differentiating} or gradient unrolling \citep{liu2021investigating} to compute $\frac{d \mathbf w^{(e)}(\btheta)}{ d \btheta}$.
In our work, we adopt the gradient unrolling method, which approximates $\mathbf w^{(e)}(\boldsymbol \theta)$ by a $K$-step gradient descent  solution, noted by $\mathbf w^{(e)}_K(\boldsymbol \theta)$ and then leverages automatic differentiation (AD) to compute the derivative from  $\mathbf w^{(e)}_K(\boldsymbol \theta)$ to the variable $\btheta$. Figure\,\ref{fig: alg_illustration} shows the working pipeline of {\ours} and its comparison to original \textsc{IRM} and {\game} methods. We use $K=1$ for the lower-level problem throughout our experiments. We refer readers to Appendix\,\ref{app: bloc_impl} for more algorithmic details. We also explore the performance of our proposed {\ours} with various regularization terms, based on the penalties used in the existing literature. We show the best performance is always achieved when the stationarity is penalized in the upper-level  (see Table\,\ref{tab: more_blo}).

%% file: sections_SL/sec_experiment.tex
\section{Experiments}
\label{sec: exp}
In this section, we begin by introducing some key experiment setups (with details in Appendix\,\ref{app: exp_setup}), and then 
empirically show the effectiveness of our proposed IRM training and evaluation improvements  over existing IRM methods across various datasets, models, and learning environments.

\subsection{Experiment Setups}
\label{sec: exp_setup}
 
\noindent\textbf{Datasets and models.}
Our experiments are conducted over $7$ datasets as  referenced and shown in  Tables\,\ref{tab: method_overview}, \ref{tab: dataset}. Among these datasets, {\CMNIST}, {\CFMNIST}, {\CM}, and {\COBJECT} are similarly curated,   mimicking the pipeline of  {\CMNIST} \citep{arjovsky2019invariant}, by introducing an environment bias parameter (\textit{e.g.}, $\beta$ for {\CMNIST} in Section\,\ref{sec: evaluation}) to customize the level of spurious correlation (as shown in Table\,\ref{tab: dataset}) in different environments. 
In the \textsc{CelebA} dataset, we choose the face attribute `{s}miling' (vs. `{n}on-smiling') as the \textit{core} feature    aimed for classification, and regard another face attribute `hair color'  (`{b}lond' vs. `{d}ark') as the source of spurious correlation imposed on the core feature. By controlling the level of spurious correlation, we then create different training/testing environments in \textsc{CelebA}. Furthermore, we   study      \textsc{PACS}  and  \textsc{VLCS} datasets, which were used to benchmark   domain generalization ability  in the real world \citep{borlino2021rethinking}. It was recently shown by \citet{gulrajani2020search} that for these datasets, \textsc{ERM} could even be better than {\irm}. Yet, we will show that our proposed \ref{eq: prob_IRM_game_consensus} is a promising domain generalization method, which outperforms all the IRM baselines and ERM in practice.
In addition, we follow \citet{arjovsky2019invariant} in adopting multi-layer perceptron (MLP) as the   model for  resolving {\CMNIST} and {\CFMNIST} problems. In the other more complex datasets, we use the ResNet-18 architecture \citep{he2016deep}.

\noindent\textbf{Baselines and implementation.} Our baselines include 7 IRM variants (Table 1) and ERM, which are implemented using their official repositories if available (see Appendix\,\ref{app: baseline}). Unless specified otherwise, our training pipeline uses the small-batch training setting. By default, we use the batch size of 1024 for {\CMNIST} and {\CFMNIST}, and 256 for other datasets. 
In Section \ref{sec:exp_results} below, we also do a thorough comparison of large-batch vs small-batch IRM training.

\begin{wraptable}{r}{.48\textwidth}
\vspace*{-7mm}
\centering
\caption{\footnotesize{Performance of existing IRM methods in large and small-batch settings. {\oracle} refers to  ERM on uncolored data, which yields the  best   prediction (supposing no  spurious correlation during training). The IRM performance is evaluated by average accuracy (`Avg Acc') and accuracy gap (`Acc Gap'), in the format  $\text{mean}{\tiny{{\pm\text{std}}}}$. A higher Avg Acc and lower Acc Gap is preferred. The theoretically optimal performance is 75\% \citep{arjovsky2019invariant}.
}}
\label{tab: bs_study}
\resizebox{.48\textwidth}{!}{%
\begin{tabular}{cc|cccc}
\toprule[1pt]
\midrule

& Dataset          
& \multicolumn{2}{c}{\CMNIST} 
& \multicolumn{2}{c}{\CFMNIST} 
\\

& Metrics(\%)      
& Avg Acc ($\uparrow$)    
& Acc Gap ($\downarrow$)  
& Avg Acc ($\uparrow$)      
& Acc Gap ($\downarrow$) \\ 
\midrule
& \oracle & 73.39\footnotesize{$\pm0.16$} & 0.32\footnotesize{$\pm0.03$} & 74.05\footnotesize{$\pm0.09$} & 0.13\footnotesize{$\pm0.04$} 
\\

& \erm
& 49.19\footnotesize{$\pm1.89$} 
& 90.72\footnotesize{$\pm2.08$}
& 49.77\footnotesize{$\pm1.71$} 
& 88.62\footnotesize{$\pm2.49$}  \\

\midrule

\parbox[t]{1mm}{\multirow{7}{1mm}{\rotatebox[origin=c]{90}{\large \textbf{Large Batch}}}}
& \irm & 67.13 \footnotesize{$\pm0.33$}& 3.43\footnotesize{$\pm0.14$}  & 67.19\footnotesize{$\pm0.22$} & 3.35\footnotesize{$\pm0.11$} 
\\

& \irmzero & 65.39 \footnotesize{$\pm0.34$}& 4.69\footnotesize{$\pm0.18$} & 66.44\footnotesize{$\pm0.28$} & 3.53\footnotesize{$\pm0.13$} 
\\

& \game  & 65.69 \footnotesize{$\pm0.42$} & 8.75\footnotesize{$\pm0.14$} & 65.91\footnotesize{$\pm0.29$} & 3.74\footnotesize{$\pm0.09$} 
\\

& \rex & 67.42 \footnotesize{$\pm0.29$} & 3.76\footnotesize{$\pm0.07$} & 67.82\footnotesize{$\pm0.31$} & 3.26\footnotesize{$\pm0.16$} 
\\

& \birm & 67.93 \footnotesize{$\pm0.31$} & 3.81\footnotesize{$\pm0.11$} & 67.75\footnotesize{$\pm0.26$} & 3.81\footnotesize{$\pm0.11$} 
\\

& \sirm & 67.72 \footnotesize{$\pm0.28$} & 3.65\footnotesize{$\pm0.08$} & 67.89\footnotesize{$\pm0.30$} & 3.12\footnotesize{$\pm0.15$} 
\\

& \fishr & 67.49\footnotesize{$\pm0.39$}  & 4.37\footnotesize{$\pm$0.10}  & 67.33\footnotesize{$\pm0.24$} & 4.49\footnotesize{$\pm0.16$} 
\\

\midrule

\parbox[t]{1mm}{\multirow{7}{1mm}{\rotatebox[origin=c]{90}{\large \textbf{ Small Batch}}}}
& \irm  
& 68.33\footnotesize{$\pm0.31$}     
& 2.04\footnotesize{$\pm0.05$}  
& 68.76\footnotesize{$\pm0.31$} 
& 1.45\footnotesize{$\pm0.09$}  \\

& \irmzero  
& 68.37\footnotesize{$\pm0.28$}     
& 1.32\footnotesize{$\pm0.09$}  
& 69.07\footnotesize{$\pm0.27$} 
& 1.36\footnotesize{$\pm0.06$}  \\

& \game 
& 67.73\footnotesize{$\pm0.24$} 
& 1.67\footnotesize{$\pm0.14$} 
& 67.49\footnotesize{$\pm0.32$} 
& 1.82\footnotesize{$\pm0.13$} \\

& \rex 
& 68.42\footnotesize{$\pm0.29$} 
& 1.65\footnotesize{$\pm0.07$} 
& 68.66\footnotesize{$\pm0.22$} 
& 1.29\footnotesize{$\pm0.08$} \\

& \birm 
& 68.71\footnotesize{$\pm0.21$} 
& 1.35\footnotesize{$\pm0.09$}  
& 68.64\footnotesize{$\pm0.32$} 
& 1.44\footnotesize{$\pm0.13$} \\

& \sirm  
& 68.81\footnotesize{$\pm0.25$}  
& 1.72\footnotesize{$\pm0.05$}  
& 68.29\footnotesize{$\pm0.22$} 
& 1.28\footnotesize{$\pm0.15$} \\

& \fishr  
& 68.69\footnotesize{$\pm0.19$}  
& 2.13\footnotesize{$\pm0.08$}  
& 68.79\footnotesize{$\pm0.17$} 
& 1.77\footnotesize{$\pm0.10$} \\
\midrule
\bottomrule[1pt]
\end{tabular}%
}
\vspace*{-13mm}
\end{wraptable}
\noindent\textbf{Evaluation setup.}
As proposed in Section \ref{sec: evaluation}, we use the multi-environment evaluation metric unless specified otherwise. To capture both the accuracy and variance of invariant predictions across multiple testing environments, the \textit{average accuracy} and the \textit{accuracy gap} (the difference of the best-case and worst-case accuracy) are measured for IRM methods.
The resulting performance is reported in the form $a\pm b$, with mean $a$ and standard deviation $b$ computed across 10 independent trials.

\subsection{Experiment Results}
\label{sec:exp_results}

\noindent\textbf{Small-batch training improves  all existing IRM methods on {\CMNIST} \& {\CFMNIST}.}
Recall from Section \ref{sec: large-batch-improvement} that all the existing IRM methods (Table\,\ref{tab: method_overview}) adopt full-batch IRM training on {\CMNIST} \& {\CFMNIST}, which raises the large-batch training problem. In \textbf{Table\,\ref{tab: bs_study}}, we conduct a thorough comparison between the originally-used full-batch IRM methods and their small-batch counterparts. In addition, we present the performance of ERM and ERM-grayscale (we call it `grayscale'), where the latter is ERM on \textit{uncolored} data. In the absence of any spurious correlation in the training set, grayscale gives the \textit{best} performance. As discussed in Section \ref{sec: evaluation} \& \ref{sec: exp_setup}, the IRM performance is measured by the average accuracy and the accuracy gap across $19$ testing environments, parameterized by the environment bias parameter $\beta \in \{ 0.05, \ldots, 0.95\}$.
We make some key observations from Table \ref{tab: bs_study}. 
\textbf{First}, small batch size helps improve \textit{all} the existing IRM methods consistently, evidenced by the $1\% \sim 3\%$ improvement in average accuracy.
\textbf{Second}, the small-batch IRM training significantly reduces the variance of invariant predictions across different testing environments, evidenced by the decreased accuracy gap. 
This implies that the small-batch IRM training can also help resolve the limitation of multi-environment evaluation for the existing IRM methods, like the sensitivity of {\game} accuracy to $\beta$ in Figure \ref{fig: motiv_eval}. 
\textbf{Third}, we observe that {\irmzero}, which does not seem to be useful in the large batch setting, becomes quite competitive with the other baselines in the small-batch setting. Thus, large-batch could suppress the IRM performance for some methods. In the rest of the experiments, we stick to the small-batch implementation of IRM training.

\begin{table}[tbh]
\centering
\caption{\footnotesize{IRM performance comparison between {{\oursTable}} and other baselines. We use ResNet-18 \citep{he2016deep} for all the datasets. The evaluation setup is consistent with Table\,\ref{tab: bs_study}, and the best performance
per-dataset is highlighted in \textbf{bold}. We present the results with the full dataset list in Table\,\ref{tab: exp_main_backup}.
}}
\label{tab: exp_main}
\resizebox{\textwidth}{!}{%
\begin{tabular}{c|cc|cc|cc|cc|cc}
\toprule[1pt]
\midrule
Algorithm & \multicolumn{2}{c}{\COBJECT} & \multicolumn{2}{c}{\CM} & \multicolumn{2}{c}{\CELEBA} & \multicolumn{2}{c}{VLCS} & \multicolumn{2}{c}{PACS} \\ 
Metrics (\%) & Avg Acc & Acc Gap & Avg Acc & Acc Gap & Avg Acc & Acc Gap & Avg Acc & Acc Gap & Avg Acc & Acc Gap \\
\midrule
\erm 
& 41.11\footnotesize{$\pm1.44$} 
& 86.43\footnotesize{$\pm2.89$}  
& 40.39\footnotesize{$\pm1.32$} 
& 85.53\footnotesize{$\pm2.33$}
& 72.38\footnotesize{$\pm 0.29$} 
& 10.73\footnotesize{$\pm 0.36$} 
& 63.23\footnotesize{$\pm 0.23$} 
& 12.39\footnotesize{$\pm 0.35$} 
& 69.95\footnotesize{$\pm 0.35$} 
& 14.32\footnotesize{$\pm 0.75$} \\
\midrule
\irm 
& 64.42\footnotesize{$\pm0.21$} 
& 4.18\footnotesize{$\pm0.29$}  
& 61.49\footnotesize{$\pm0.29$} 
& 7.17\footnotesize{$\pm0.33$}  
& 72.49\footnotesize{$\pm 0.38$} 
& 10.15\footnotesize{$\pm 0.27$} 
& 62.72\footnotesize{$\pm 0.29$} 
& 12.74\footnotesize{$\pm 0.27$} 
& 68.93\footnotesize{$\pm 0.33$} 
& 14.99\footnotesize{$\pm 0.51$} 
\\
\irmzero 
& 62.39\footnotesize{$\pm0.25$} 
& 5.36\footnotesize{$\pm0.31$}  
& 60.14\footnotesize{$\pm0.18$} 
& 8.83\footnotesize{$\pm0.39$}  
& 72.42\footnotesize{$\pm 0.35$} 
& 10.43\footnotesize{$\pm 0.38$} 
& 62.59\footnotesize{$\pm 0.32$} 
& 12.99\footnotesize{$\pm 0.36$} 
& 68.72\footnotesize{$\pm 0.29$} 
& 15.29\footnotesize{$\pm 0.71$} 
\\
\game 
& 62.88\footnotesize{$\pm0.34$} 
& 5.59\footnotesize{$\pm0.28$}  
& 60.44\footnotesize{$\pm0.31$} 
& 6.72\footnotesize{$\pm0.41$}  
& 72.18\footnotesize{$\pm 0.44$} 
& 12.32\footnotesize{$\pm 0.41$} 
& 62.31\footnotesize{$\pm 0.38$} 
& 13.37\footnotesize{$\pm 0.62$} 
& 68.12\footnotesize{$\pm 0.22$} 
& 15.77\footnotesize{$\pm 0.66$} 
\\
\rex 
& 63.37\footnotesize{$\pm0.35$} 
& 5.42\footnotesize{$\pm0.31$}  
& 62.32\footnotesize{$\pm0.24$} 
& 5.55\footnotesize{$\pm0.32$}  
& 72.34\footnotesize{$\pm 0.26$} 
& 10.31\footnotesize{$\pm 0.23$} 
& 63.19\footnotesize{$\pm 0.31$} 
& 12.87\footnotesize{$\pm 0.31$} 
& 69.43\footnotesize{$\pm 0.34$} 
& 15.31\footnotesize{$\pm 0.67$} 
\\
\birm 
& 65.11\footnotesize{$\pm0.27$} 
& \textbf{3.31}\footnotesize{$\pm0.22$}  
& 62.99\footnotesize{$\pm0.35$} 
& 5.23\footnotesize{$\pm0.36$}  
& 72.93\footnotesize{$\pm 0.28$} 
& 9.92\footnotesize{$\pm 0.33$} 
& 63.33\footnotesize{$\pm 0.40$} 
& 12.13\footnotesize{$\pm 0.23$} 
& 69.34\footnotesize{$\pm 0.25$} 
& 15.76\footnotesize{$\pm 0.49$} 
\\
\sirm 
& 64.97\footnotesize{$\pm0.39$} 
& 3.97\footnotesize{$\pm0.25$}  
& 62.16\footnotesize{$\pm0.29$} 
& \textbf{4.14}\footnotesize{$\pm0.31$}  
& 72.42\footnotesize{$\pm 0.33$} 
& 9.79\footnotesize{$\pm 0.21$} 
& 62.86\footnotesize{$\pm 0.26$} 
& 12.79\footnotesize{$\pm 0.35$} 
& 69.52\footnotesize{$\pm 0.39$} 
& 15.81\footnotesize{$\pm 0.82$} 
\\ 
{\fishr} 
& 64.07\footnotesize{$\pm0.23$} 
& 4.41\footnotesize{$\pm0.29$} 
& 61.79\footnotesize{$\pm0.25$} 
& 5.55\footnotesize{$\pm0.21$} 
& 72.89\footnotesize{$\pm 0.25$} 
& 9.42\footnotesize{$\pm 0.32$} 
& 63.44\footnotesize{$\pm 0.37$} 
& 11.93\footnotesize{$\pm 0.42$} 
& 70.21\footnotesize{$\pm 0.22$} 
& \textbf{14.52}\footnotesize{$\pm 0.43$} 
\\
\rowcolor[gray]{.8}
\oursTable 
& \textbf{65.97}\footnotesize{$\pm0.33$} 
& 4.10\footnotesize{$\pm0.36$}  
& \textbf{63.69}\footnotesize{$\pm0.32$} 
& 4.89\footnotesize{$\pm0.36$}  
& \textbf{73.35}\footnotesize{$\pm 0.32$} 
& \textbf{8.79}\footnotesize{$\pm 0.21$} 
& \textbf{63.62}\footnotesize{$\pm 0.35$} 
& \textbf{11.55}\footnotesize{$\pm 0.32$} 
& \textbf{70.31}\footnotesize{$\pm 0.21$} 
& 14.73\footnotesize{$\pm 0.65$} 
\\
\midrule
\bottomrule[1pt]
\end{tabular}%
}
\vspace*{-5mm}
\end{table}

\noindent\textbf{{\ours} outperforms IRM baselines in various datasets.}
Next, \textbf{Table\,\ref{tab: exp_main}} demonstrates the effectiveness of our proposed {\ours} approach versus ERM and existing IRM baselines across all the 7 datasets listed in Table\,\ref{tab: dataset}.  
Evidently, {\ours} yields a higher average accuracy compared to all the baselines, together with the smallest accuracy gap in most cases. 
Additionally, we observe that \textsc{CelebA}, \textsc{PACS}  and  \textsc{VLCS} are much more challenging datasets for capturing invariance through IRM, as evidenced by the small performance gap between ERM and  IRM methods.
In particular, all the IRM methods, except {\fishr} and {\ours}, could even be worse than {\erm} on \textsc{PACS}  and  \textsc{VLCS}. Here, we echo and extend the findings of \citet[Section~4.3]{krueger2021out}. However, we also show that {\ours} is a quite competitive IRM variant when applied to realistic domain generalization datasets. We also highlight that
the {\CELEBA} experiment is newly constructed and performed in our work for invariance evaluation. Like \textsc{PACS}  and  \textsc{VLCS}, this experiment also shows that {\erm} is a strong baseline, and among IRM-based methods, {\ours} is the best-performing, both in terms of accuracy and variance of invariant predictions. 

\begin{wrapfigure}{r}{.32\textwidth}
\vspace*{-5mm}
    \centering
    \includegraphics[width=.3\textwidth]{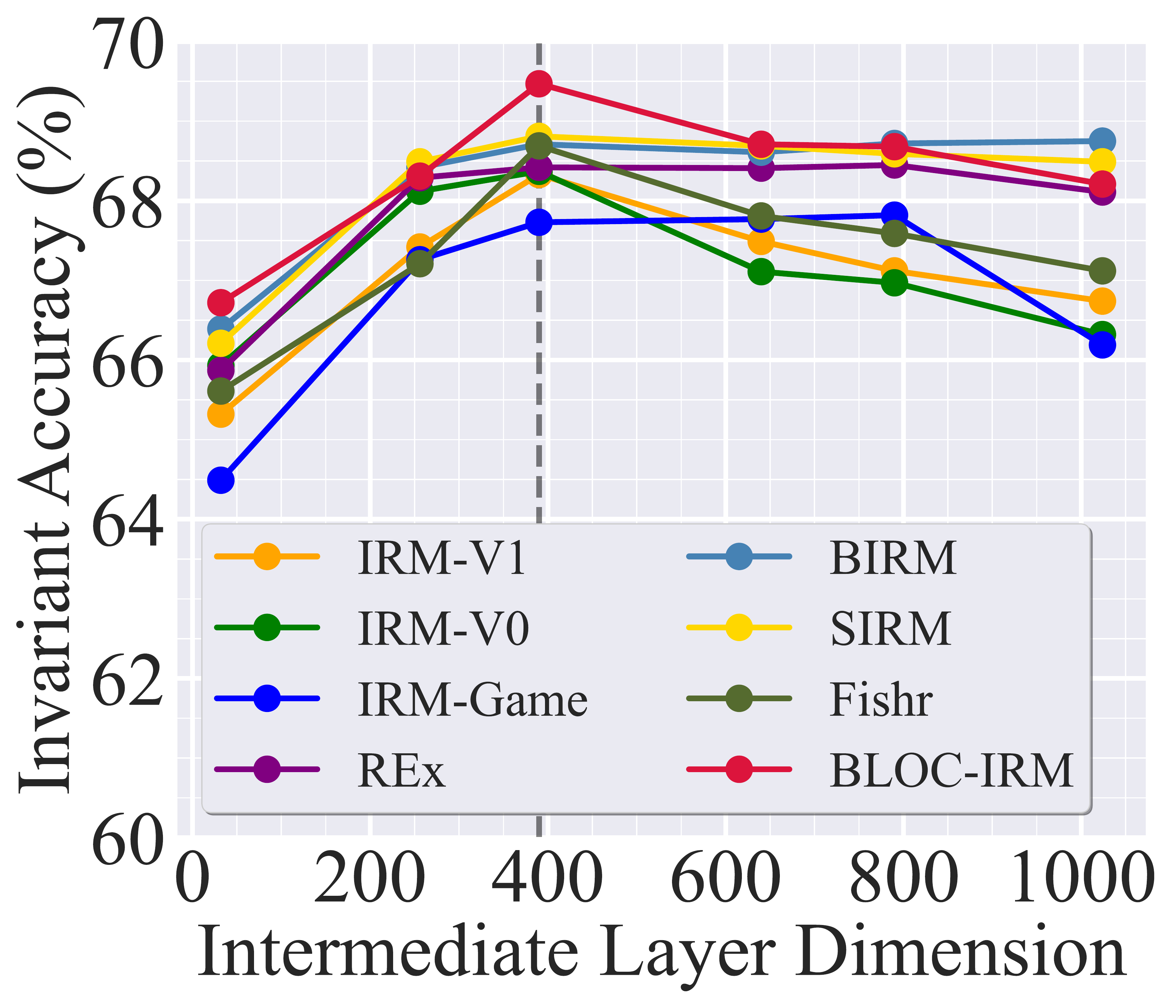}
    \vspace*{-1em}
    \caption{\footnotesize{IRM performance on {\CMNIST} against the layer dimension in MLP. 
    The  dotted line represents the default dimension ($d=390$) used in the literature. The invariant prediction accuracy is presented via the dot line (mean). The results are based on 10 independent trials and we report the variance in Figure\,\ref{fig: exp_model_size_with_err}.
    }}
    \label{fig: exp_model_size}
    \vspace*{-2mm}
\end{wrapfigure}

\noindent\textbf{IRM against model size and training environment variation.} Furthermore, we investigate the effect of model size and training environment diversity on the IRM performance. The recent works \citep{lin2022bayesian, zhou2022sparse} have empirically shown that {\irm} may suffer a significant performance loss when trained over large-sized neural network models, and thus  developed   {\birm} and {\sirm} approaches as   advancements of {\irm}. 
Inspired by these works, \textbf{Figure\,\ref{fig: exp_model_size}} presents the sensitivity of invariant  prediction  to model size for different IRM methods on {\CMNIST}. Here the model size is  controlled by the   dimension of the intermediate layer (denoted by $d$) in MLP, and the default dimension is {$d = 390$} (\textit{i.e.}, the vertical dotted line in Figure \ref{fig: exp_model_size}), which was used in \citep{arjovsky2019invariant} and followed in the subsequent literature. As we can see, when $d > 390$, nearly all the studied IRM methods (including {\ours}) suffer a performance drop. 
Yet, as $d \geq 800$, from the perspective of prediction accuracy and model resilience together, the top-3 best IRM methods with model size resilience are {\birm}, {\sirm}, and {\ours}, although we did not intentionally design {\ours} to resist performance degradation against model size. 

We also show more experiment results in the Appendix. In \textbf{Table\,\ref{tab: exp_training_env}}, we study IRM with different numbers of training environment configurations and observe the consistent improvement of {\ours} over other baselines. 
In \textbf{Table~\ref{tab: imbalance}} we show that the performance of invariant prediction degrades, if additional covariate shifts (class, digit, and color imbalances on {\CMNIST}) are imposed on the training environments following  \citet[Section~4.1]{krueger2021out} and also demonstrate that {\ours} maintains the accuracy improvement over baselines with each variation. In \textbf{Table\,\ref{tab: exp_fail_case}}, we compare the performance of different methods in the failure cases of IRM pointed out by \citep{kamath2021does} and show the consistent improvement brought by {\ours}.

%% file: appendix.tex
\appendix
\onecolumn
\setcounter{section}{0}

\setcounter{section}{0}
\setcounter{figure}{0}

\makeatletter 
\renewcommand{\thefigure}{A\arabic{figure}}
\renewcommand{\theHfigure}{A\arabic{figure}}
\renewcommand{\thetable}{A\arabic{table}}
\renewcommand{\theHtable}{A\arabic{table}}

\makeatother
\setcounter{table}{0}

\setcounter{mylemma}{0}
\renewcommand{\themylemma}{A\arabic{mylemma}}
\setcounter{algorithm}{0}
\renewcommand{\thealgorithm}{A\arabic{algorithm}}
\setcounter{equation}{0}
\renewcommand{\theequation}{A\arabic{equation}}

\begin{center}
    \Large \textsc{Appendix}
\end{center}

\section{Dataset Selection}
\label{app: dataset}


Compared to existing work, we expand the dataset types for evaluating the performance of different IRM methods (see Table\,\ref{tab: dataset}).
In addition to the most commonly-used benchmark  datasets {\CMNIST} \citep{arjovsky2019invariant} and {\CFMNIST} \citep{ahuja2020invariant}, we also consider the datasets {\CM} \citep{lin2021empirical, shah2020pitfalls} and {\COBJECT} \citep{ahmed2020systematic, zhang2021can}, which impose artificial spurious correlations, MNIST digit number and object color, into the original CIFAR-10 and COCO Detection datasets, respectively.
Furthermore, we consider other three  real-world datasets \textsc{CelebA} \citep{liu2015faceattributes}, \textsc{PACS} \citep{li2017deeper} and \textsc{VLCS} \citep{torralba2011unbiased}, without imposing artificial spurious correlations. 
Notably, \textsc{CelebA} was first formalized and introduced to benchmark IRM performance. The recent work \citep{gulrajani2020search} showed that when carefully implemented, ERM could outperform {\irm} in  \textsc{PACS} and \textsc{VLCS}. Thus, we regard them as challenging datasets to capture invariance.

For {\COBJECT} dataset, we strictly follow the setting adopted in \citep{lin2022bayesian} to generate the spurious features. For {\CM} we use the class ``bird'' and ``plane'' in the dataset \texttt{CIFAR} as the invariant feature, while the digit ``0'' and ``1'' in \texttt{MNIST} as the spurious correlation. 

{\CELEBA} dataset is, for the first time, introduced to measure IRM performance. We select the attribute ``Smiling'' as the invariant label and use the attribute ``Hair Color'' (blond and black hair) to create a spurious correlation in each environment.

\section{Implementation Details}
\label{app: impl}

\subsection{Details on Large-batch Optimization Enhancements}
\label{app: large_impl}

\ding{70} \textbf{\irmLSGD}: We first integrate large-batch SGD (LSGD) with IRM.  Following \citep{goyal2017accurate}, we make two main modifications: (1) scaling up learning rate linearly with batch size, and (2) prepending a warm-up optimization phase to IRM training. We call the  LSGD-baked IRM variant {\irmLSGD}.

\ding{70} \textbf{{\irmLALR}}: Next, we adopt layerwise adaptive learning rate (LALR) in IRM training.  Following \citep{you2019large}, we advance the learning rate scheduler by assigning each layer of a neural network-based prediction model with an \textit{adaptive} learning rate (\textit{i.e.}, proportional to the norm of updated model weights per layer). More specifically, the model parameter update rule becomes:
\begin{align}
\small
    \btheta_{t+1, i} = \btheta_{t, i} - \frac{\tau(\|\btheta_{t,i}\|_2^2) \cdot  \eta_t}{\|\mathbf{u}_{t,i}\|_2^2} \mathbf{u}_{t,i},
\end{align}%
where $\btheta_{t,i}$ denotes the $i$-th layer of the model parameters at iteration $t$, and $\mathbf{u}_{t,i}$ represents the first-order gradient of the corresponding  layer-wise model parameters. We use $\tau(\|\btheta_{t,i}\|_2^2 = \text{min}\{\text{max}\{\|\btheta_{t,i}\|_2^2,c_l\}, c_{u}\})$ as the scaling factor of the adaptive learning rate $\frac{\eta_t}{\|\mathbf{u}_{t,i}\|}$. We use $c_l=0$ and $c_u=1$ in our experiments.

\ding{70} \textbf{{\irmSAM}}: Lastly, we leverage  sharpness-aware minimization (SAM)  to simultaneously minimize the IRM loss and the loss sharpness. The latter is achieved by  explicitly penalizing the worst-case training loss of model weights when facing  small weight perturbations. This yields a wide minimum within a flat loss landscape. More specifically, the sharpness-aware loss can be formulated as:
\begin{align}
    \small
    \min_{\btheta} \ell^{\text{SAM}}(\btheta), \quad \text{where} \qquad \ell^{\text{SAM}}(\btheta) = \max_{\|\boldsymbol\epsilon\|_2^2\leq \rho} \ell(\btheta + \boldsymbol\epsilon),
\end{align}%
where the parameter perturbation $\boldsymbol\epsilon$ is subject to the perturbation constraint $\|\boldsymbol\epsilon\|_2^2 \leq \rho$. When applied to IRM, we replace the per-environment training loss with the SAM loss, and adopt the $\rho = 0.001$.

\subsection{{\ours} Implementation}
\label{app: bloc_impl}
As described in Section\,\ref{sec: consensus}, the {\ours} algorithm solves the IRM problem with two optimization levels. We use $1$-step gradient descent to get the lower-level solution. We retain the gradient graph in PyTorch to enable auto differentiation. We assign each of the classification head $\{\bw^{(e)}\}$ a separate optimizer and use the same learning rate as the feature extractor $\btheta$. For {\CMNIST} and {\CFMNIST}, we adopt a learning rate of $2\times 10^{-3}$ and use the Adam \citep{kingma2014adam} optimizer. As for other datasets, we use the multi-step learning rate scheduler with an initial learning rate of $0.1$, which is consistent with other baselines. We adopt the same penalty weight of $10^6$ as {\irm} and {\irmzero}.


\begin{algorithm}[H]
        \caption{{\ours}}
        \begin{algorithmic}[1]
        \State \textbf{Initialization}: Training data $\{\bx^{(e)}\}$ from $N$ environments, Model feature extractor $\btheta_0$, and $N$ model classification heads $\{\bw^{(e)}_0\}$, learning rate $\{\eta_t\}$ series, penalty weight $\{\gamma_t\}$ series.
        
        \For{Step $t=0,1,\ldots,$}
            \State \textbf{Lower-level:} update classification head for each environment:
            \begin{align}
            \small
                \forall e \in \Envtr, \qquad \tilde{\bw}^{(e)}_{t+1} = \bw^{(e)}_{t} - \eta_t \left . \frac{d \ell^{(e)}(\bw \odot \btheta)}{d\bw} \right |_{\btheta = \btheta_{t}, \bw = \bw^{(e)}_{t}}
            \end{align}%
            
            \State \textbf{Consensus projection:}
            $\forall e \in \Envtr, \bw^{(e)}_{t+1} = \bw^*_{t+1} = \frac{1}{N}\sum_{e\in \Envtr} \tilde{\bw}^{(e)}_{t+1}$
            
            \State \textbf{Upper-level:} update feature extractor with stationary penalty:
            
            \begin{align}
            \small
                \btheta_{t+1} = \btheta_t - \eta_t \sum_{e\in \Envtr} \frac{d}{d\btheta}
                \left(\left . {\ell^{(e)}(\bw \odot \btheta)}
                + 
                \gamma_t \| \nabla_{\mathbf w} \Le (\mathbf w \circ \boldsymbol \theta) \|_2^2
                \right) \right |_{\btheta = \btheta_t, \bw=\bw^*_{t+1}}
            \end{align}
            
        \EndFor
        \end{algorithmic}
        \label{alg: bloc}
\end{algorithm}






\section{Experimentation}

\subsection{Environment Setup}
\label{app: exp_setup}
As proposed in Section \ref{sec: evaluation}, we use the multi-environment evaluation metric unless specified otherwise. To capture both the accuracy and variance of invariant predictions across multiple testing environments, the \textit{average accuracy} and the \textit{accuracy gap} (the difference between the best-case and worst-case accuracy) are evaluated for IRM methods.

Specifically, for the {\CMNIST}, {\CFMNIST}, {\COBJECT}, {\CM}, and {\CELEBA} dataset, we manually create 19 test environments with uniformly sampled bias parameter $\beta \in \{0.05, 0.1, \dots, 0.95\}$, where the environment bias parameter $\beta$ controls the spurious correlation (see Section\,\ref{sec: evaluation} for more details).

For VLCS and PACS datasets, the training and test sets have 4 environments, namely \{art painting, cartoon, sketch, photo\} and \{CALTECH, LABELME, PASCAL, SUN\} respectively. We use the first three environments as the training environments, while we use the test set of all four environments to form our proposed multi-environment invariance evaluation system.

\subsection{Baselines}
\label{app: baseline}
For each baseline method, we follow its official PyTorch repository except {\IRMGame} and {\sirm}. We translate the TensorFlow-based original code base of {\IRMGame} to PyTorch. As one of the latest IRM advancements, the official code of {\sirm} is not yet publicly available. Therefore, we reproduce {\sirm} in PyTorch.

In particular, for {\CMNIST} and {\CFMNIST}, we stick to the original hyper-parameters for the large-batch setting and tune the hyper-parameters of each method, including the penalty weight, number of warm-up epochs, and learning rate for the small batch setting. 

In particular, for the large-batch setting, we use the penalty weight of $10^6$, 190 warm-up epochs, and 500 epochs in total, as suggested by the original {\irm} and inherited by its variants. For the small-batch setting, we adopt the same penalty weight $10^6$. Further, we found that the warm-up phase could be shortened without sacrificing accuracy. Therefore, we use 50 warm-up epochs and total 200 epochs for all the methods.

For other datasets, we adopt the batch size of 128 and use ResNet-18 as the default model architecture. We train for 200 epochs. We adopt the step-wise learning rate scheduler with an initial learning rate of $0.1$. The learning rate decays by $0.1$ at the $100$th and $150$th epochs.

\subsection{Additional Experiment Results}
\label{app: results}

\paragraph{The influence of batch size with all the baselines.}

\begin{figure}[htb]
    \centering
    \includegraphics[width=.35\textwidth]{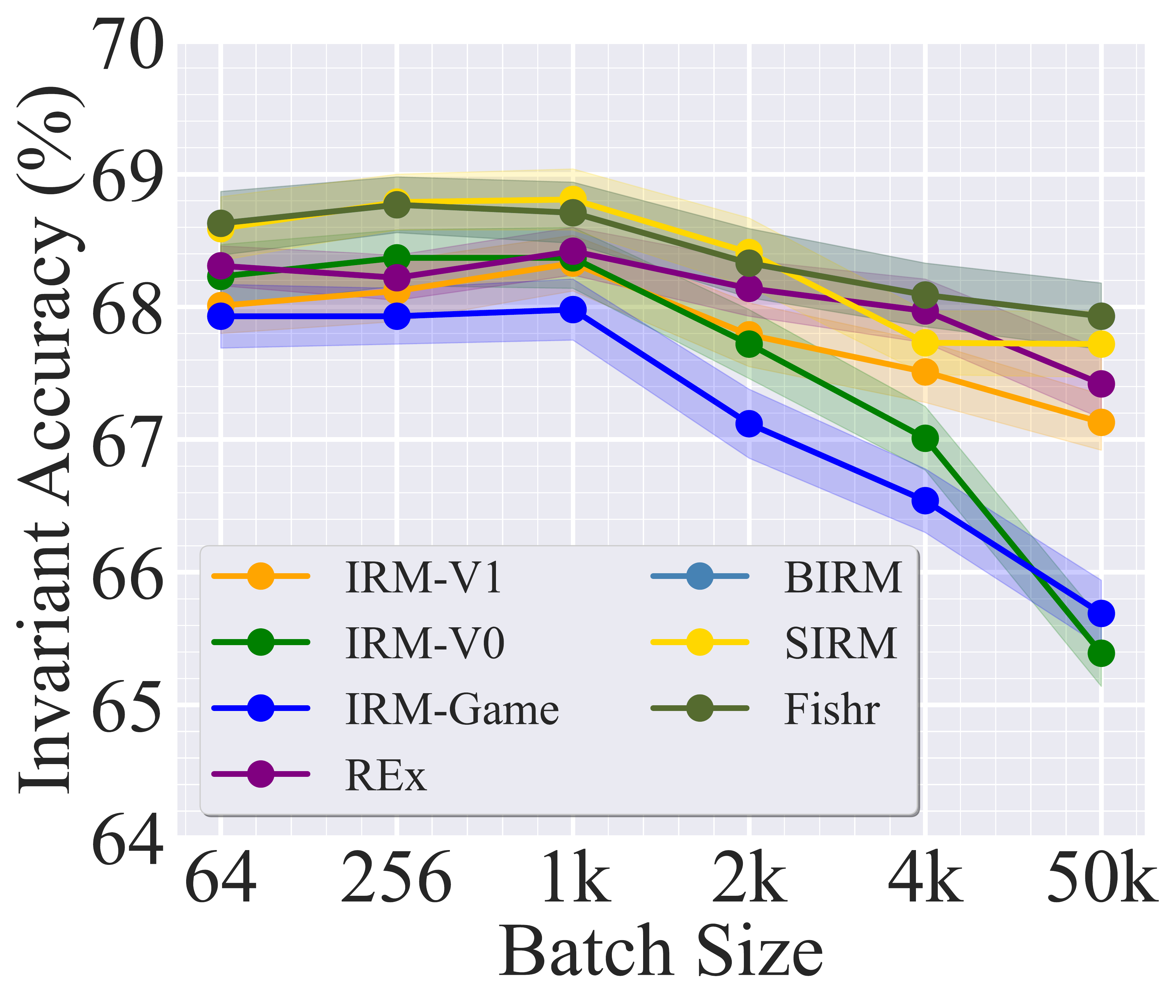}
    \caption{\footnotesize{The performance of all the baselines in this work trained with different batch sizes on {\CMNIST} dataset. The full data batch-size is 50k. The invariant accuracy corresponds to the average accuracy evaluated based on the diversified environments-based evaluation metric.}}
    \label{fig: bs_full}
\end{figure}

We show in Figure\,\ref{fig: bs_full} the influence of training batch size on the performance of different methods. We observe in Figure\,\ref{fig: bs_full}, as in Figure\,\ref{fig: acc_bs}, that full batch setting does not achieve the best performance, and the use of mini-batch (stochastic gradient descent) indeed improves performance.

\paragraph{Loss landscapes of {\irm} with different batch sizes.}

\begin{figure}[htb]
\centering
\resizebox{.8\textwidth}{!}{
\begin{tabular}{cc}
\includegraphics[width=0.3\textwidth]{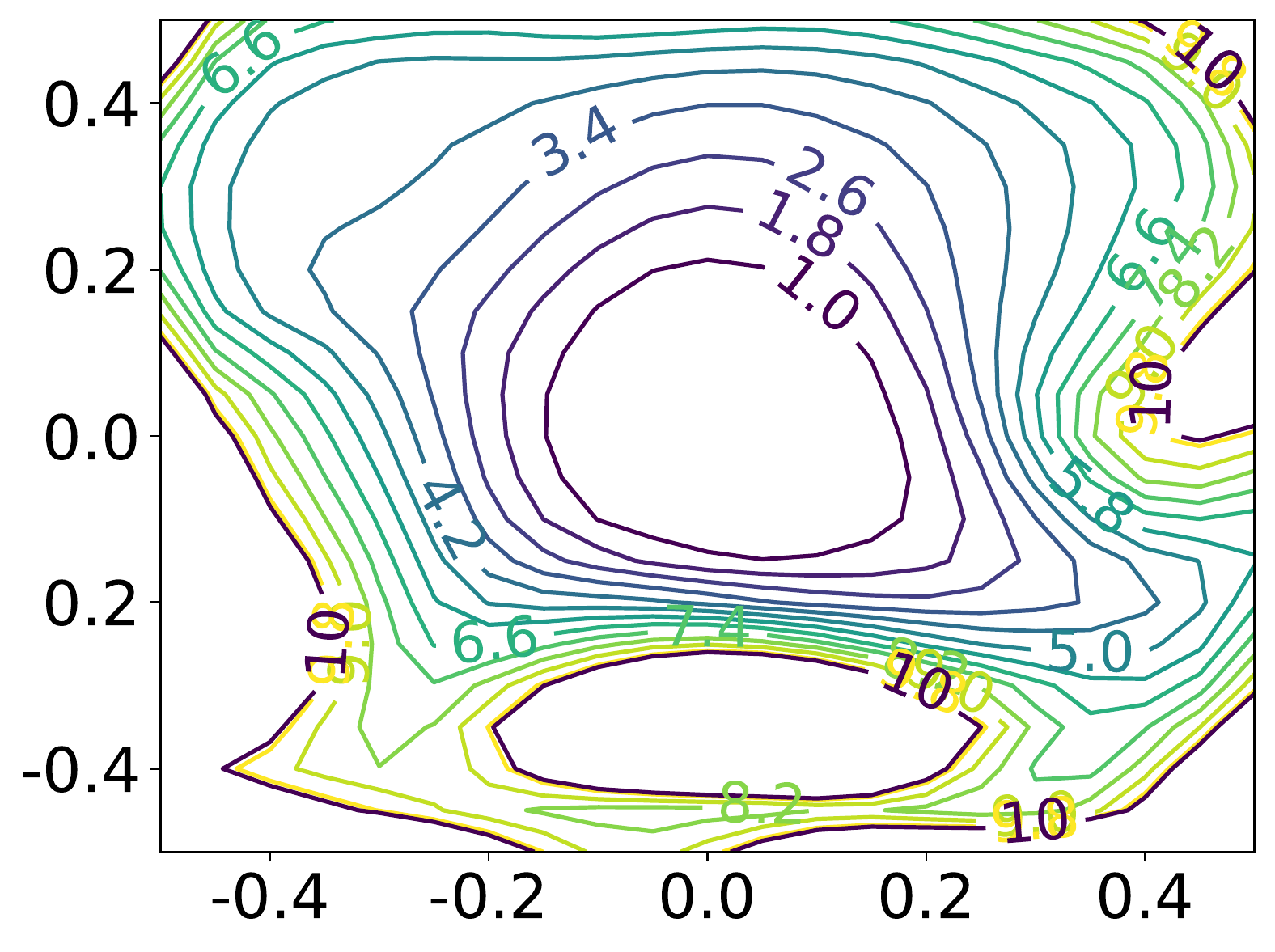}
& \includegraphics[width=0.3\textwidth]{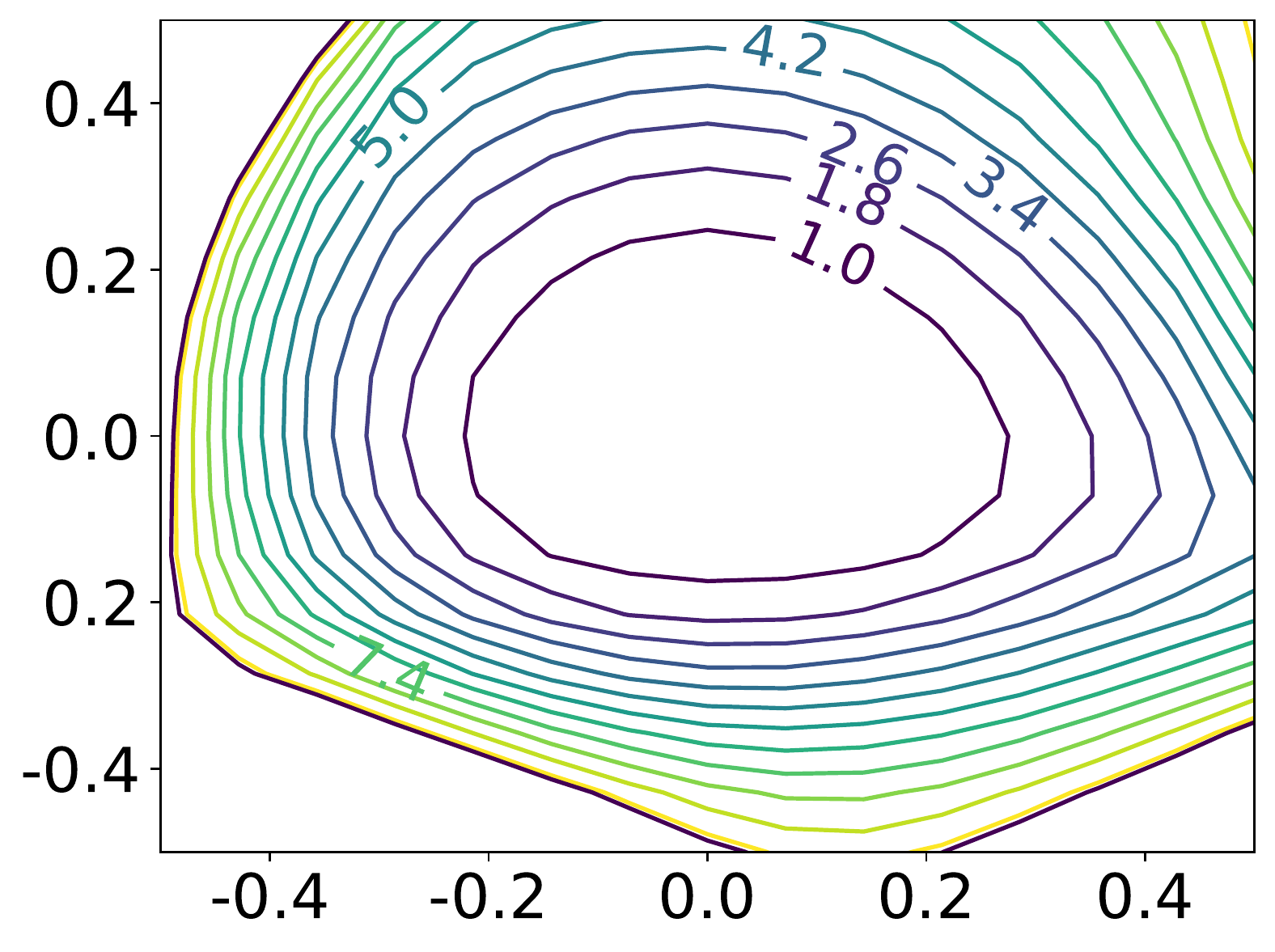} \\
{\scriptsize{(A) Large Batch}}  & {\scriptsize	{(B) Small Batch}}
\end{tabular}
}
\caption{\footnotesize{The loss landscapes of invariant prediction models acquired by (A)
large-batch {\irm} training with 50k batch size and (B) small-batch training with 1k batch size. The 2D loss landscape visualization is realized using  the tool in \citep{li2018visualizing}. The $x$ and $y$ axes   represent the linear interpolation coefficients  over two directional vectors originated from the converged local optima. Here the numbers on the contour denote the loss values over   test data. 
}}
\label{fig: trajactory_bs}
\end{figure} 
We plot the loss landscapes of the models trained with {\irm} on {\CMNIST} using large (full) and small batch in Figure\,\ref{fig: trajactory_bs}. Using small batch training, {\irm} (Fig.\,\ref{fig: trajactory_bs}B) converges to a smooth neighborhood of a local optima. This also corresponds to a flatter loss landscape than the landscape of the large-batch training (Figure\,\ref{fig: trajactory_bs}(A)). The loss landscapes demonstrate consistent results as other experiments discussed in Section\,\ref{sec: large-batch-improvement}.


\paragraph{Training trajectory with {\ours} with and without stationary loss.}

\begin{figure}[htb]
    \centering
    \includegraphics[width=.39\textwidth]{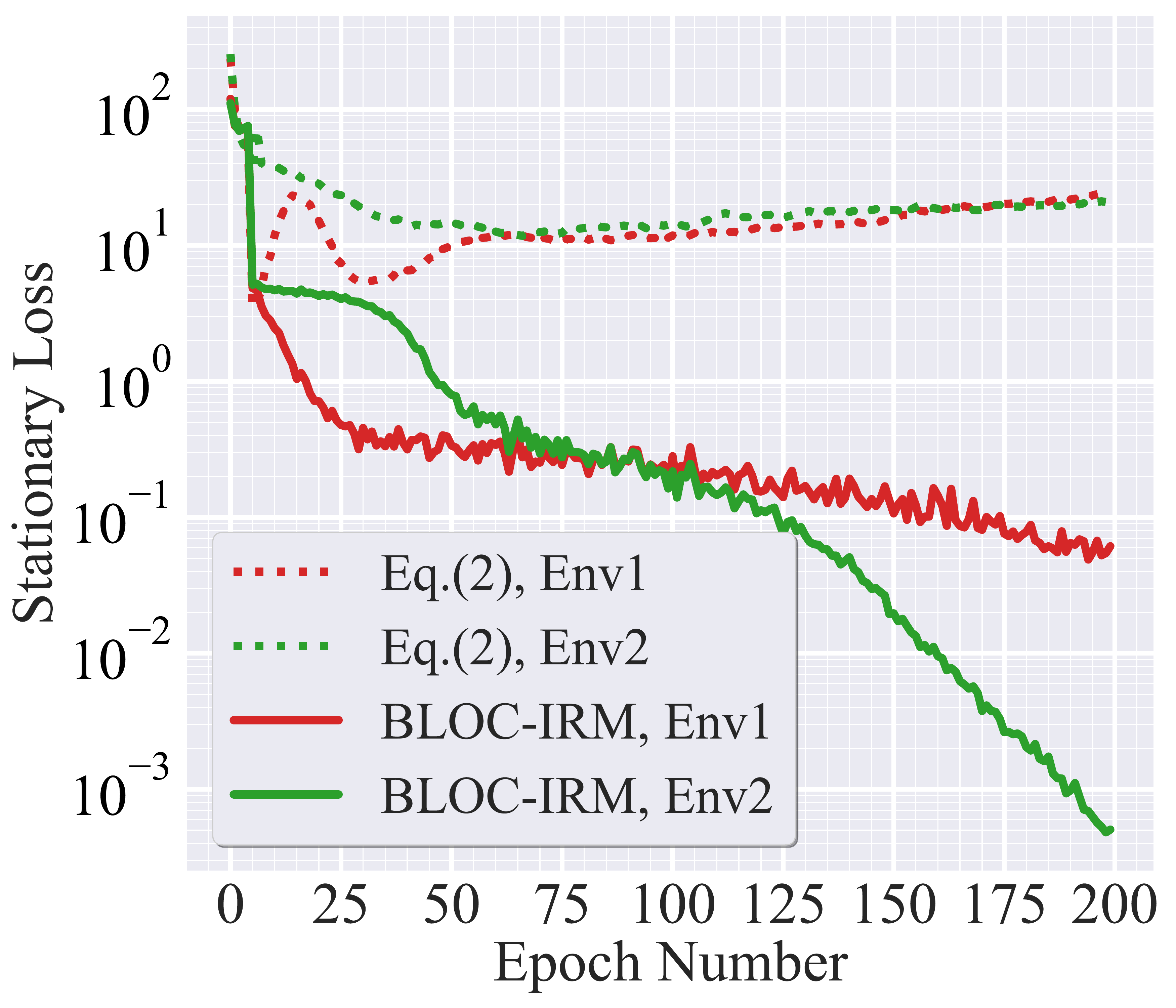}
     \vspace*{-1mm}
    \caption{\footnotesize{The per-environment training trajectory for the stationarity loss of \eqref{eq: prob_IRM_game_invariance} and \eqref{eq: prob_IRM_game_consensus} on {\CMNIST}. The training setting is the same as Figure\,\ref{fig: motiv_eval}. The algorithmic details can be found in Appendix\,\ref{app: impl}.}}
    \label{fig: stationary_loss}
\end{figure}

In Figure\,\ref{fig: stationary_loss}, we plot the per-environment training trajectory of stationary loss when solving \eqref{eq: prob_IRM_game_invariance} and \eqref{eq: prob_IRM_game_consensus} on {\CMNIST}. For \eqref{eq: prob_IRM_game_consensus} we use the regularization term $\lambda=10^6$, which is aligned with the penalty coefficient used in {\irm}. As we can see, without the stationarity regularization, the stationary loss remains at a high level for both environments (the dotted curves). Notably, the lower-level stationary can be reached fast with the stationarity penalty, as shown in the solid curves.

\paragraph{Performance of all the methods with full dataset list.}

We show in Table\,\ref{tab: exp_main_backup} the results of all the methods on the seven datasets we studied. To be more specific, in Table\,\ref{tab: exp_main_backup}, we append the results of {\CMNIST} and {\CFMNIST} into Table\,\ref{tab: exp_main} as a whole. As we can see, our methods outperforms other baselines in all the datasets in terms of average accuracy, and stands top in most cases in terms of the accuracy gap. 

\begin{table}[htb]
\centering
\caption{\footnotesize{IRM performance comparison between our proposed {{\oursTable}} method and other baselines under the full list of datasets. We use MLP for {\CMNIST} and {\CFMNIST}, and ResNet-18 \citep{he2016deep} for the rest datasets. The evaluation setup is consistent with Table\,\ref{tab: bs_study}, and the best performance per-evaluation metric and per-dataset is highlighted in \textbf{bold}.
}}
\label{tab: exp_main_backup}
\resizebox{\textwidth}{!}{%
\begin{tabular}{c|cc|cc|cc|cc|cc|cc|cc}
\toprule[1pt]
\midrule
Algorithm & \multicolumn{2}{c}{\CMNIST} & \multicolumn{2}{c}{\CFMNIST} & \multicolumn{2}{c}{\COBJECT} & \multicolumn{2}{c}{\CM} & \multicolumn{2}{c}{\CELEBA} & \multicolumn{2}{c}{VLCS} & \multicolumn{2}{c}{PACS} \\ 
Metrics (\%) & Avg Acc & Acc Gap & Avg Acc & Acc Gap & Avg Acc & Acc Gap & Avg Acc & Acc Gap & Avg Acc & Acc Gap & Avg Acc & Acc Gap & Avg Acc & Acc Gap \\
\midrule
\erm
& 49.19\footnotesize{$\pm1.89$} 
& 90.72\footnotesize{$\pm2.08$}
& 49.77\footnotesize{$\pm1.71$} 
& 88.62\footnotesize{$\pm2.49$}
& 41.11\footnotesize{$\pm1.44$} 
& 86.43\footnotesize{$\pm2.89$}  
& 40.39\footnotesize{$\pm1.32$} 
& 85.53\footnotesize{$\pm2.33$}
& 72.38\footnotesize{$\pm 0.29$} 
& 10.73\footnotesize{$\pm 0.36$} 
& 63.23\footnotesize{$\pm 0.23$} 
& 12.39\footnotesize{$\pm 0.35$} 
& 69.95\footnotesize{$\pm 0.35$} 
& 14.32\footnotesize{$\pm 0.75$} \\
\midrule
\irm
& 68.33\footnotesize{$\pm0.31$}     
& 2.04\footnotesize{$\pm0.05$}  
& 68.76\footnotesize{$\pm0.31$} 
& 1.45\footnotesize{$\pm0.09$}
& 64.42\footnotesize{$\pm0.21$} 
& 4.18\footnotesize{$\pm0.29$}  
& 61.49\footnotesize{$\pm0.29$} 
& 7.17\footnotesize{$\pm0.33$}  
& 72.49\footnotesize{$\pm 0.38$} 
& 10.15\footnotesize{$\pm 0.27$} 
& 62.72\footnotesize{$\pm 0.29$} 
& 12.74\footnotesize{$\pm 0.27$} 
& 68.93\footnotesize{$\pm 0.33$} 
& 14.99\footnotesize{$\pm 0.51$} 
\\
\irmzero
& 68.37\footnotesize{$\pm0.28$}     
& 1.32\footnotesize{$\pm0.09$}  
& 69.07\footnotesize{$\pm0.27$} 
& 1.36\footnotesize{$\pm0.06$}
& 62.39\footnotesize{$\pm0.25$} 
& 5.36\footnotesize{$\pm0.31$}  
& 60.14\footnotesize{$\pm0.18$} 
& 8.83\footnotesize{$\pm0.39$}  
& 72.42\footnotesize{$\pm 0.35$} 
& 10.43\footnotesize{$\pm 0.38$} 
& 62.59\footnotesize{$\pm 0.32$} 
& 12.99\footnotesize{$\pm 0.36$} 
& 68.72\footnotesize{$\pm 0.29$} 
& 15.29\footnotesize{$\pm 0.71$} 
\\
\game
& 67.73\footnotesize{$\pm0.24$} 
& 1.67\footnotesize{$\pm0.14$} 
& 67.49\footnotesize{$\pm0.32$} 
& 1.82\footnotesize{$\pm0.13$}
& 62.88\footnotesize{$\pm0.34$} 
& 5.59\footnotesize{$\pm0.28$}  
& 60.44\footnotesize{$\pm0.31$} 
& 6.72\footnotesize{$\pm0.41$}  
& 72.18\footnotesize{$\pm 0.44$} 
& 12.32\footnotesize{$\pm 0.41$} 
& 62.31\footnotesize{$\pm 0.38$} 
& 13.37\footnotesize{$\pm 0.62$} 
& 68.12\footnotesize{$\pm 0.22$} 
& 15.77\footnotesize{$\pm 0.66$} 
\\
\rex
& 68.42\footnotesize{$\pm0.29$} 
& 1.65\footnotesize{$\pm0.07$} 
& 68.66\footnotesize{$\pm0.22$} 
& 1.29\footnotesize{$\pm0.08$}
& 63.37\footnotesize{$\pm0.35$} 
& 5.42\footnotesize{$\pm0.31$}  
& 62.32\footnotesize{$\pm0.24$} 
& 5.55\footnotesize{$\pm0.32$}  
& 72.34\footnotesize{$\pm 0.26$} 
& 10.31\footnotesize{$\pm 0.23$} 
& 63.19\footnotesize{$\pm 0.31$} 
& 12.87\footnotesize{$\pm 0.31$} 
& 69.43\footnotesize{$\pm 0.34$} 
& 15.31\footnotesize{$\pm 0.67$} 
\\
\birm
& 68.71\footnotesize{$\pm0.21$} 
& 1.35\footnotesize{$\pm0.09$}  
& 68.64\footnotesize{$\pm0.32$} 
& 1.44\footnotesize{$\pm0.13$}
& 65.11\footnotesize{$\pm0.27$} 
& \textbf{3.31}\footnotesize{$\pm0.22$}  
& 62.99\footnotesize{$\pm0.35$} 
& 5.23\footnotesize{$\pm0.36$}  
& 72.93\footnotesize{$\pm 0.28$} 
& 9.92\footnotesize{$\pm 0.33$} 
& 63.33\footnotesize{$\pm 0.40$} 
& 12.13\footnotesize{$\pm 0.23$} 
& 69.34\footnotesize{$\pm 0.25$} 
& 15.76\footnotesize{$\pm 0.49$} 
\\
\sirm
& 68.81\footnotesize{$\pm0.25$}  
& 1.72\footnotesize{$\pm0.05$}  
& 68.29\footnotesize{$\pm0.22$} 
& 1.28\footnotesize{$\pm0.15$}
& 64.97\footnotesize{$\pm0.39$} 
& 3.97\footnotesize{$\pm0.25$}  
& 62.16\footnotesize{$\pm0.29$} 
& \textbf{4.14}\footnotesize{$\pm0.31$}  
& 72.42\footnotesize{$\pm 0.33$} 
& 9.79\footnotesize{$\pm 0.21$} 
& 62.86\footnotesize{$\pm 0.26$} 
& 12.79\footnotesize{$\pm 0.35$} 
& 69.52\footnotesize{$\pm 0.39$} 
& 15.81\footnotesize{$\pm 0.82$} 
\\ 
{\fishr}
& 68.69\footnotesize{$\pm0.19$}  
& 2.13\footnotesize{$\pm0.08$}  
& 68.79\footnotesize{$\pm0.17$} 
& 1.77\footnotesize{$\pm0.10$}
& 64.07\footnotesize{$\pm0.23$} 
& 4.41\footnotesize{$\pm0.29$} 
& 61.79\footnotesize{$\pm0.25$} 
& 5.55\footnotesize{$\pm0.21$} 
& 72.89\footnotesize{$\pm 0.25$} 
& 9.42\footnotesize{$\pm 0.32$} 
& 63.44\footnotesize{$\pm 0.37$} 
& 11.93\footnotesize{$\pm 0.42$} 
& 70.21\footnotesize{$\pm 0.22$} 
& \textbf{14.52}\footnotesize{$\pm 0.43$} 
\\
\rowcolor[gray]{.8}
\oursTable
& \cellcolor[gray]{.8}\textbf{69.47}\footnotesize{$\pm0.24$} 
& \cellcolor[gray]{.8}\textbf{1.04}\footnotesize{$\pm0.07$} 
& \cellcolor[gray]{.8}\textbf{69.43}\footnotesize{$\pm0.21$} 
& \cellcolor[gray]{.8}\textbf{1.14}\footnotesize{$\pm0.11$} 
& \textbf{65.97}\footnotesize{$\pm0.33$} 
& 4.10\footnotesize{$\pm0.36$}  
& \textbf{63.69}\footnotesize{$\pm0.32$} 
& 4.89\footnotesize{$\pm0.36$}  
& \textbf{73.35}\footnotesize{$\pm 0.32$} 
& \textbf{8.79}\footnotesize{$\pm 0.21$} 
& \textbf{63.62}\footnotesize{$\pm 0.35$} 
& \textbf{11.55}\footnotesize{$\pm 0.32$} 
& \textbf{70.31}\footnotesize{$\pm 0.21$} 
& 14.73\footnotesize{$\pm 0.65$} 
\\
\midrule
\bottomrule[1pt]
\end{tabular}%
}
\end{table}

\paragraph{Experiment on different model sizes.}

\begin{figure}[htb]
    \centering
    \includegraphics[width=.39\textwidth]{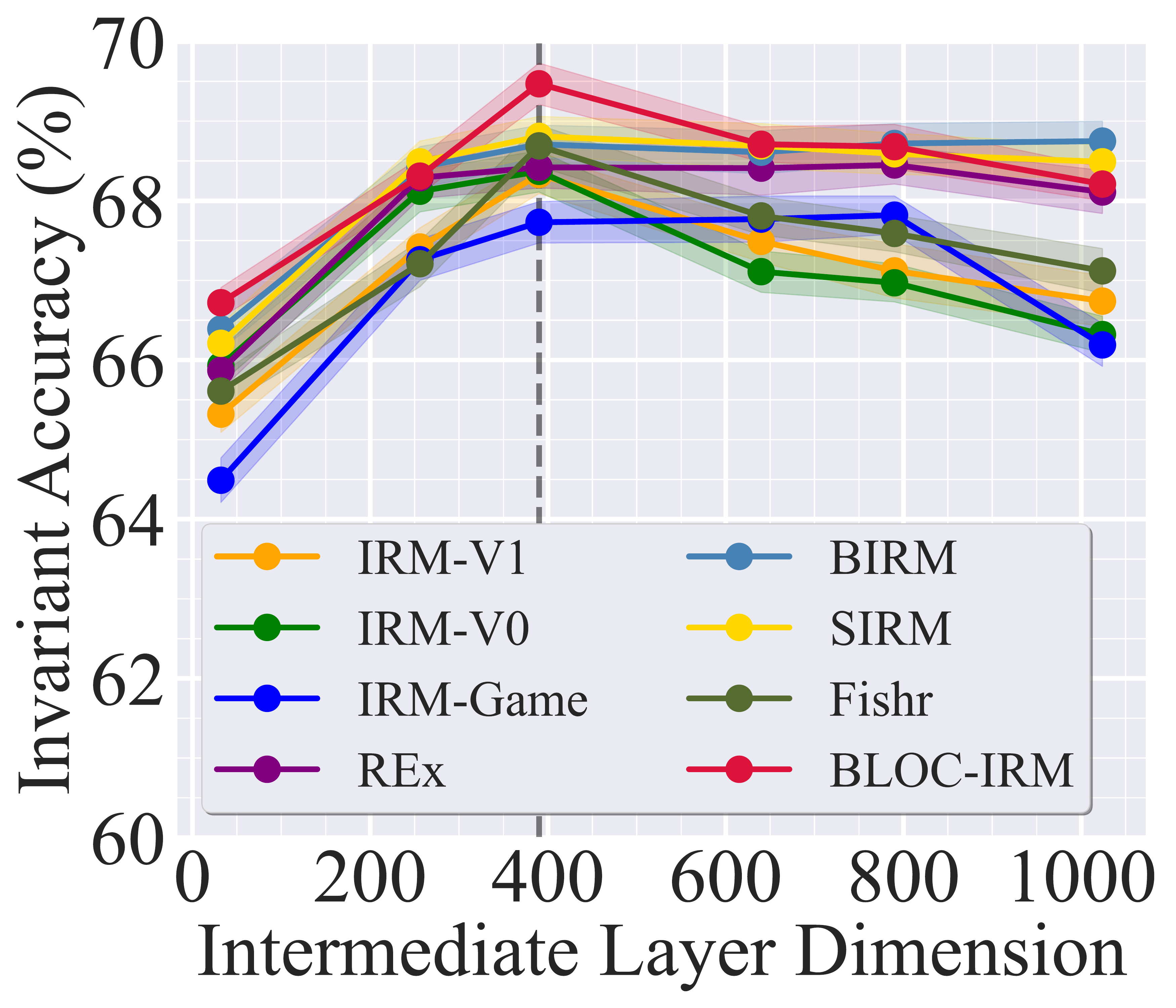}
    \vspace*{-1mm}
    \caption{\footnotesize{IRM performance on {\CMNIST} against the dimension of the intermediate layer in MLP. 
    The  dotted line represents the default dimension ($d=390$) used in the literature. The invariant prediction accuracy is presented via the dot line (mean) and shaded area (standard deviation) over $10$ random trials.
    }}
    \label{fig: exp_model_size_with_err}
\end{figure}

We show in Figure\,\ref{fig: exp_model_size_with_err} the influence of the increasing model size on the performance of different baselines considered in this work. Compared to Figure\,\ref{fig: exp_model_size}, we report additional standard deviation of the 10 independent trials in Figure\,\ref{fig: exp_model_size_with_err}.

\clearpage

\paragraph{Experiment with different training environments.}

In Table\,\ref{tab: exp_training_env}, we show the performance of all the methods in more complex training environments, such as more training environments and more skewed environment bias parameter $\beta$. As we can see, {\ours} outperforms other baselines.

\begin{table}[htb]
\centering
\caption{\footnotesize{Performance under different training environments in {\CMNIST}.}}
\label{tab: exp_training_env}
\resizebox{.6\textwidth}{!}{%
\begin{tabular}{c|cccc}
\toprule[1pt]
\midrule
Environment      & \multicolumn{2}{c}{$p_{\mathtt{tr}} \in \{0.1, 0.15\}$} & \multicolumn{2}{c}{$p_{\mathtt{tr}} \in \{0.1, 0.15, 0.2\}$} \\ 
Metrics (\%)          & Avg Acc        & Acc Gap        & Avg Acc           & Acc Gap           \\ 
\midrule
\optim          & 75.00               & 0.00               & 75.00               & 0.00                   \\
\oracle & 73.82\footnotesize{$\pm0.11$} & 0.37\footnotesize{$\pm0.05$} & 73.97\footnotesize{$\pm0.14$} & 0.29\footnotesize{$\pm0.08$}\\ 
\erm              & 49.21\footnotesize{$\pm0.79$} & 91.88\footnotesize{$\pm3.31$} & 49.03\footnotesize{$\pm0.93$} & 92.17\footnotesize{$\pm3.04$}\\
\midrule
\irm & 67.36\footnotesize{$\pm0.31$} & 2.77\footnotesize{$\pm0.15$} & 67.11\footnotesize{$\pm0.34$} & 2.42\footnotesize{$\pm0.12$} \\
\irmzero & 67.01\footnotesize{$\pm0.42$} & 2.85\footnotesize{$\pm0.18$} & 66.71\footnotesize{$\pm0.42$} & 2.36\footnotesize{$\pm0.19$} \\
\game & 66.39\footnotesize{$\pm0.72$} & 4.47\footnotesize{$\pm0.61$} & 65.93\footnotesize{$\pm0.53$} & 4.25\footnotesize{$\pm0.84$} \\
\rex & 66.82\footnotesize{$\pm0.44$} & 2.59\footnotesize{$\pm0.11$} & 67.14\footnotesize{$\pm0.38$} & 2.16\footnotesize{$\pm0.13$} \\
\birm & 67.35\footnotesize{$\pm0.39$} & 2.65\footnotesize{$\pm0.10$} & 68.05\footnotesize{$\pm0.43$} & \textbf{1.99}\footnotesize{$\pm0.07$} \\
\sirm & 67.12\footnotesize{$\pm0.53$} & 2.33\footnotesize{$\pm0.18$} & 67.72\footnotesize{$\pm0.41$} & 2.11\footnotesize{$\pm0.19$} \\
\fishr & 67.22\footnotesize{$\pm0.43$} & 2.44\footnotesize{$\pm0.15$} & 67.32\footnotesize{$\pm0.39$} & 2.59\footnotesize{$\pm0.15$} \\
\rowcolor[gray]{.8}
BLO-IRM  & \textbf{68.72}\footnotesize{$\pm0.41$} & \textbf{2.19}\footnotesize{$\pm0.15$} & \textbf{68.89}\footnotesize{$\pm0.31$} & 2.39\footnotesize{$\pm0.09$} \\
\midrule
\bottomrule[1pt]
\end{tabular}%
}
\end{table}

\paragraph{{\ours} with different regularizations.}

Based on the penalty terms used in the existing IRM variants, we explore the performance of our proposed {\ours} with various regularization, including the ones used in {\irm} (\textsc{BLOC-IRM-v1}), {\rex} (\textsc{BLOC-IRM-rex}), and {\fishr} (\textsc{BLOC-IRM-Fishr}). We conduct experiments on three different datasets and the results are shown in Table\,\ref{tab: more_blo}. It is obvious that the best performance is always achieved when the per-environment stationarity is penalized in the upper-level. This is not surprising since without an explicit promotion of stationarity, other forms of penalties do not guarantee the BLO algorithm to achieve an optimal solution.

\begin{table}[htb]
\centering
\caption{\footnotesize{The performance of {\ours} with different regularization terms. Three datasets are studied and the latest baseline {\sirm} is listed as reference for comparison. The best performance per-evaluation metric and per-dataset is highlighted in \textbf{bold}.}}
\label{tab: more_blo}
\resizebox{.8\textwidth}{!}{%
\begin{tabular}{c|cccccc}
\toprule[1pt]
\midrule
Dataset & \multicolumn{2}{c}{\CMNIST} & \multicolumn{2}{c}{\COBJECT} & \multicolumn{2}{c}{\CM} \\
Metrics & Avg Acc & Acc Gap & Avg Acc & Acc Gap & Avg Acc & Acc Gap \\ 
\midrule
\sirm & 68.81\footnotesize{$\pm0.25$} & 1.72\footnotesize{$\pm0.05$} & 64.97\footnotesize{$\pm0.39$} & \textbf{3.97}\footnotesize{$\pm0.25$} & 62.87\footnotesize{$\pm0.29$} & \textbf{4.14}\footnotesize{$\pm0.31$} \\
\midrule
\textsc{BLOC-IRM} & \textbf{69.47}\footnotesize{$\pm0.24$} & \textbf{1.04}\footnotesize{$\pm0.07$} & \textbf{65.97}\footnotesize{$\pm0.33$} & 4.10\footnotesize{$\pm0.36$} & \textbf{63.69}\footnotesize{$\pm0.32$} & 4.89\footnotesize{$\pm0.36$} \\
\textsc{BLOC-IRM-v1} & 67.14\footnotesize{$\pm0.24$} & 4.33\footnotesize{$\pm0.83$} & 63.38\footnotesize{$\pm0.29$} & 6.31\footnotesize{$\pm0.51$} & 61.13\footnotesize{$\pm0.51$} &  6.71\footnotesize{$\pm0.41$} \\
\textsc{BLOC-IRM-REx} & 62.71\footnotesize{$\pm0.21$} & 8.74\footnotesize{$\pm1.21$} & 60.31\footnotesize{$\pm0.33$} & 7.62\footnotesize{$\pm0.66$} & 59.39\footnotesize{$\pm0.55$} & 7.89\footnotesize{$\pm0.45$} \\
\textsc{BLOC-IRM-Fishr} &
63.25\footnotesize{$\pm0.16$} & 7.12\footnotesize{$\pm0.39$} & 61.17\footnotesize{$\pm0.34$} & 6.98.\footnotesize{$\pm0.45$} & 60.86\footnotesize{$\pm0.51$} & 6.63\footnotesize{$\pm0.30$} \\
\midrule
\bottomrule[1pt]
\end{tabular}%
}
\end{table}

\paragraph{Performance comparison of different methods with additional covariate shifts.} Besides the sensitivity check on model size, \textbf{Table~\ref{tab: imbalance}} examines the resilience of IRM to variations in the training environment. This study is motivated by \citet{krueger2021out}, who empirically showed that the performance of invariant prediction degrades if additional covariate shifts are imposed on the training environments. Thus, we  present the IRM performance on {\CMNIST} by introducing class, digit, and color imbalances, following  \citet[Section~4.1]{krueger2021out}. Compared with Table~\ref{tab: bs_study}, IRM suffers a greater performance loss in Table~\ref{tab: imbalance}, in the presence of  training environment variations. However, the proposed {\ours} maintains the accuracy improvement over baselines with each variation. In Table\,\ref{tab: exp_training_env}, we also study IRM with different numbers of training environments and observe the consistent improvement of {\ours} over other baselines. 

\begin{table}[thb]
\centering
\caption{\footnotesize{IRM performance 
 on {\CMNIST} and {\CFMNIST} with training environment variations in terms of  class, digit and color imbalances. The best IRM performance per-evaluation
metric and per-variation source is highlighted in \textbf{bold}.} }
\label{tab: imbalance}
\resizebox{\textwidth}{!}{%
\begin{tabular}{c|cccccc|cccccc}
\toprule[1pt]
\midrule
Dataset & \multicolumn{6}{c}{\CMNIST} & \multicolumn{6}{c}{\CFMNIST} \\
\midrule
Variation  & \multicolumn{2}{c}{Class Imbalance} & \multicolumn{2}{c}{Digit Imbalance} & \multicolumn{2}{c}{Color Imbalance} & \multicolumn{2}{c}{Class Imbalance} & \multicolumn{2}{c}{Digit Imbalance} & \multicolumn{2}{c}{Color Imbalance} \\ 
Metrics (\%) & Avg Acc  & Acc Gap & Avg Acc & Acc Gap         & Avg Acc & Acc Gap & Avg Acc  & Acc Gap & Avg Acc & Acc Gap         & Avg Acc & Acc Gap \\ 
\midrule

\oracle 
& 71.23\footnotesize{$\pm0.18$} 
& 2.76\footnotesize{$\pm0.11$}
& 70.31\footnotesize{$\pm0.21$}
& 2.79\footnotesize{$\pm0.15$}
& 72.29\footnotesize{$\pm0.16$}
& 2.88\footnotesize{$\pm0.14$}
& 70.15\footnotesize{$\pm0.21$}
& 2.29\footnotesize{$\pm0.12$}                
& 69.92\footnotesize{$\pm0.15$}                  
& 2.72\footnotesize{$\pm0.21$}                
& 73.31\footnotesize{$\pm0.11$}                   
& 1.17\footnotesize{$\pm0.23$} 
\\

\erm 
& 43.72\footnotesize{$\pm1.01$}
& 92.76\footnotesize{$\pm1.45$}
& 45.89\footnotesize{$\pm2.82$}
& 91.65\footnotesize{$\pm1.86$}
& 46.19\footnotesize{$\pm2.88$}
& 90.88\footnotesize{$\pm1.69$}
& 41.72\footnotesize{$\pm1.98$} 
& 93.37\footnotesize{$\pm2.15$} 
& 42.39\footnotesize{$\pm2.39$} 
& 92.23\footnotesize{$\pm2.72$} 
& 45.89\footnotesize{$\pm0.27$} 
& 91.31\footnotesize{$\pm2.27$}
\\

\midrule
\irm 
& 65.39\footnotesize{$\pm0.22$}
& 4.44\footnotesize{$\pm0.29$}
& 64.89\footnotesize{$\pm0.26$}
& 4.19\footnotesize{$\pm0.44$}
& 66.12\footnotesize{$\pm0.25$}
& 3.31\footnotesize{$\pm0.29$}
& 62.49\footnotesize{$\pm0.33$} 
& 4.93\footnotesize{$\pm0.45$} 
& 61.88\footnotesize{$\pm0.23$} 
& 5.54\footnotesize{$\pm0.39$} 
& 64.39\footnotesize{$\pm0.44$} 
& 3.79\footnotesize{$\pm0.33$} 
\\

\irmzero 
& 65.01\footnotesize{$\pm0.28$}
& 4.29\footnotesize{$\pm0.33$}
& 65.13\footnotesize{$\pm0.25$}
& 3.87\footnotesize{$\pm0.28$}
& 66.72\footnotesize{$\pm0.25$}
& 3.01\footnotesize{$\pm0.44$}
& 62.78\footnotesize{$\pm0.48$} 
& 5.33\footnotesize{$\pm0.47$} 
& 61.62\footnotesize{$\pm0.29$} 
& 5.29\footnotesize{$\pm0.41$} 
& 64.93\footnotesize{$\pm0.27$} 
& 3.28\footnotesize{$\pm0.31$} 
\\

\game 
& 62.21\footnotesize{$\pm0.42$}
& 6.45\footnotesize{$\pm0.35$}
& 62.10\footnotesize{$\pm0.35$}
& 6.72\footnotesize{$\pm0.44$}
& 61.82\footnotesize{$\pm0.65$}
& 7.78\footnotesize{$\pm0.55$}
&60.73\footnotesize{$\pm0.84$} 
& 6.24\footnotesize{$\pm0.43$} 
& 60.79\footnotesize{$\pm0.45$} 
& 6.47\footnotesize{$\pm0.82$} 
&64.32\footnotesize{$\pm0.42$} 
&5.73\footnotesize{$\pm0.31$} 
\\

\rex 
& \textbf{66.45}\footnotesize{$\pm0.25$}
& 3.39\footnotesize{$\pm0.28$}
& 66.23\footnotesize{$\pm0.43$}
& \textbf{3.21}\footnotesize{$\pm0.20$}
& 66.99\footnotesize{$\pm0.42$}
& 3.32\footnotesize{$\pm0.27$}
& 64.89\footnotesize{$\pm0.36$} 
& 5.78\footnotesize{$\pm0.53$} 
& 63.95\footnotesize{$\pm0.25$} 
& 4.73\footnotesize{$\pm0.62$}                
& 65.87\footnotesize{$\pm0.42$} 
& 4.30\footnotesize{$\pm0.42$} 
\\

\birm 
& 65.73\footnotesize{$\pm0.25$}
& 4.11\footnotesize{$\pm0.31$}
& 65.73\footnotesize{$\pm0.88$}
& 4.49\footnotesize{$\pm0.67$}
& 66.72\footnotesize{$\pm0.24$}
& 3.47\footnotesize{$\pm0.25$}
& 64.39\footnotesize{$\pm0.34$} 
& 4.47\footnotesize{$\pm0.39$} 
& 63.24\footnotesize{$\pm0.39$} 
& \textbf{4.54}\footnotesize{$\pm0.42$} 
& 65.08\footnotesize{$\pm0.31$} 
& 3.80\footnotesize{$\pm0.29$}
\\ 

\sirm 
& 65.32\footnotesize{$\pm0.39$}
& 4.92\footnotesize{$\pm0.22$}
& 64.44\footnotesize{$\pm0.36$}
& 4.85\footnotesize{$\pm0.33$}
& 66.03\footnotesize{$\pm0.32$}
& \textbf{2.85}\footnotesize{$\pm0.19$}
& 64.32\footnotesize{$\pm0.51$} 
& 4.15\footnotesize{$\pm0.36$} 
& 62.97\footnotesize{$\pm0.35$} 
& 5.75\footnotesize{$\pm0.52$}                
& 64.72\footnotesize{$\pm0.46$} 
& 3.99\footnotesize{$\pm0.39$}
\\

\fishr 
& 66.13\footnotesize{$\pm0.28$}
& 3.99\footnotesize{$\pm0.32$}
& 65.87\footnotesize{$\pm0.42$}
& 3.72\footnotesize{$\pm0.41$}
& 65.48\footnotesize{$\pm0.21$}
& 4.49\footnotesize{$\pm0.31$}
& 63.62\footnotesize{$\pm0.53$} 
& 5.59\footnotesize{$\pm0.35$} 
& 62.47\footnotesize{$\pm0.26$} 
& 5.72\footnotesize{$\pm0.33$} 
& 65.13\footnotesize{$\pm0.32$} 
& 4.44\footnotesize{$\pm0.21$}
\\

\rowcolor[gray]{.8}
\oursTable
& 66.32\footnotesize{$\pm0.27$}
& \textbf{3.11}\footnotesize{$\pm0.22$}
& \textbf{66.41}\footnotesize{$\pm0.29$}
& 3.32\footnotesize{$\pm0.25$}
& \textbf{67.25}\footnotesize{$\pm0.24$}
& 3.72\footnotesize{$\pm0.27$}
& \textbf{65.99}\footnotesize{$\pm0.31$}                  
& \textbf{3.97}\footnotesize{$\pm0.43$}                
& \textbf{65.13}\footnotesize{$\pm0.31$}                  
& 5.11\footnotesize{$\pm0.45$}                
& \textbf{66.79}\footnotesize{$\pm0.26$}                   
& \textbf{3.72}\footnotesize{$\pm0.36$}                
\\ 

\midrule
\bottomrule[1pt]
\end{tabular}%
}
\end{table}

\paragraph{Exploration on the failure cases of previous IRM methods.} Some papers either theoretically \citep{rosenfeld2020risks} or empirically \citep{kamath2021does} pointed out that the original {\irm} method could fail in certain circumstances, due to the fact that the regularization term used in {\irm} heavily relies on the “linear predictor” assumption. Regarding this issue, we first bring to attention that the {\ours} formulation does not require the predictors to be linear, since we adopt the regularization in the form of {\irmzero} in the upper-level objective, not {\irm}. To justify our argument, we repeat the experiments in \citep{kamath2021does}, which points out a specific scenario using the {\CMNIST} dataset where {\irm} fails. 
\begin{wraptable}{r}{0.32\linewidth}
\caption{\footnotesize{
Performance comparisons on {\CMNIST} among ERM, {\irm}, and {\ours} in the scenarios where IRM-variants failed following \citep{kamath2021does}.}}
\label{tab: exp_fail_case}
\resizebox{.3\textwidth}{!}{%
\begin{tabular}{c|cc}
\toprule[1pt]
Method  & Avg. Acc. & Acc. Gap \\
\midrule
ERM     & 83.09     & 13.79    \\
{\irm}  & 76.89     & 27.68    \\
{\ours} & 84.22     & 11.01   \\
\bottomrule[1pt]
\end{tabular}
}
\end{wraptable}
More specifically, the models are trained in the training environments $(\alpha, \beta) = (0.1, 0.2)$ and $(0.1, 0.25)$, and evaluated in the test environment $(0.1, 0.9)$. Note that denotes the label flipping rate and represents the environment bias parameter. The results are shown in the Table\,\ref{tab: exp_fail_case}. As we can see, {\irm} is clearly worse than ERM as it achieves much lower average accuracy and higher accuracy gap. However, {\ours} outperforms ERM by obtaining high average accuracy and lower accuracy gap. This result shows that {\ours} seems promising to address the empirical IRM challenge discovered in \citep{kamath2021does}.
In the meantime, we also acknowledge that BLOC-IRM is not perfect since the advantage achieved by {\ours} over ERM is not strong enough. However, we stress that the main contribution of {\ours} does not lie in solving the failure cases of {\irm}, but to fix the issue of IRM-Game that resorts to a predictor ensemble to make the invariant prediction, which deviates from the spirit of acquiring invariant predictors in the original IRM paradigm.

\begin{figure}[htb]
    \centering
    \includegraphics[width=.35\linewidth]{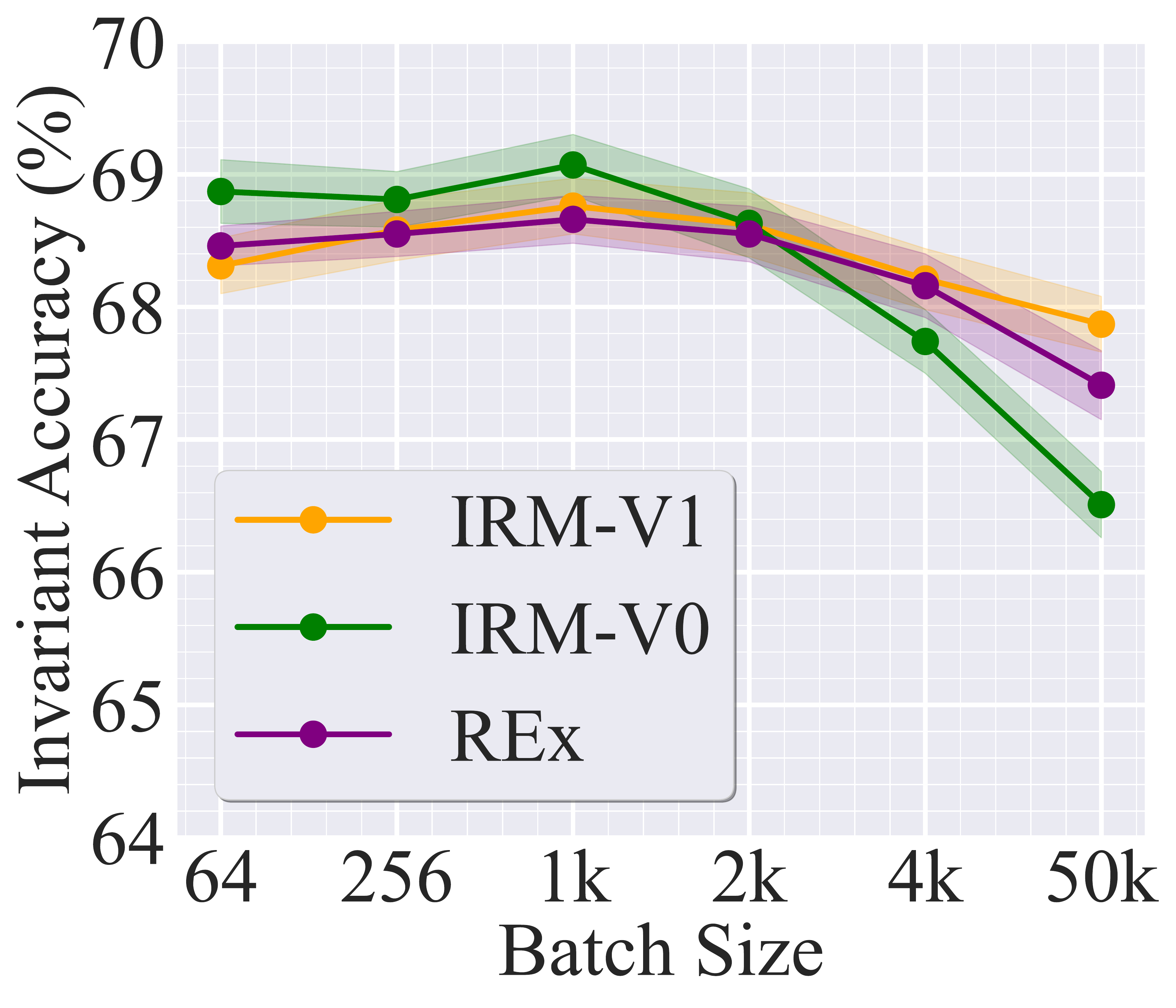}
    \caption{\footnotesize{The performance of three IRM methods ({\irm}, {\irmzero}, and {\rex}) vs. batch size under {\CFMNIST}. }}
    \label{fig: acc_bs_cfmnist}
\end{figure}

\paragraph{A similar curve to Figure\,\ref{fig: acc_bs} on {\CFMNIST}.} We show the results for {\CFMNIST} similar to Figure\,\ref{fig: acc_bs} in Figure\,\ref{fig: acc_bs_cfmnist}and the conclusion does not change much. As mentioned before, the large-batch training setup was typically used for IRM training over the {\CMNIST} and {\CFMNIST} datasets.

%% file: main.bbl
\begin{thebibliography}{62}
\providecommand{\natexlab}[1]{#1}
\providecommand{\url}[1]{\texttt{#1}}
\expandafter\ifx\csname urlstyle\endcsname\relax
  \providecommand{\doi}[1]{doi: #1}\else
  \providecommand{\doi}{doi: \begingroup \urlstyle{rm}\Url}\fi

\bibitem[Ahmed et~al.(2020)Ahmed, Bengio, van Seijen, and
  Courville]{ahmed2020systematic}
Faruk Ahmed, Yoshua Bengio, Harm van Seijen, and Aaron Courville.
\newblock Systematic generalisation with group invariant predictions.
\newblock In \emph{International Conference on Learning Representations}, 2020.

\bibitem[Ahuja et~al.(2020)Ahuja, Shanmugam, Varshney, and
  Dhurandhar]{ahuja2020invariant}
Kartik Ahuja, Karthikeyan Shanmugam, Kush Varshney, and Amit Dhurandhar.
\newblock Invariant risk minimization games.
\newblock In \emph{International Conference on Machine Learning}, pp.\
  145--155. PMLR, 2020.

\bibitem[Ahuja et~al.(2021)Ahuja, Wang, Dhurandhar, Shanmugam, and
  Varshney]{ahuja2020empirical}
Kartik Ahuja, Jun Wang, Amit Dhurandhar, Karthikeyan Shanmugam, and Kush~R
  Varshney.
\newblock Empirical or invariant risk minimization? a sample complexity
  perspective.
\newblock In \emph{International Conference on Learning Representations}, 2021.

\bibitem[Arjovsky et~al.(2019)Arjovsky, Bottou, Gulrajani, and
  Lopez-Paz]{arjovsky2019invariant}
Martin Arjovsky, L{\'e}on Bottou, Ishaan Gulrajani, and David Lopez-Paz.
\newblock Invariant risk minimization.
\newblock \emph{arXiv preprint arXiv:1907.02893}, 2019.

\bibitem[Balaji et~al.(2018)Balaji, Sankaranarayanan, and
  Chellappa]{balaji2018metareg}
Yogesh Balaji, Swami Sankaranarayanan, and Rama Chellappa.
\newblock Metareg: Towards domain generalization using meta-regularization.
\newblock \emph{Advances in neural information processing systems}, 31, 2018.

\bibitem[Beery et~al.(2018)Beery, Van~Horn, and Perona]{beery2018recognition}
Sara Beery, Grant Van~Horn, and Pietro Perona.
\newblock Recognition in terra incognita.
\newblock In \emph{Proceedings of the European conference on computer vision
  (ECCV)}, pp.\  456--473, 2018.

\bibitem[Borlino et~al.(2021)Borlino, D'Innocente, and
  Tommasi]{borlino2021rethinking}
Francesco~Cappio Borlino, Antonio D'Innocente, and Tatiana Tommasi.
\newblock Rethinking domain generalization baselines.
\newblock In \emph{2020 25th International Conference on Pattern Recognition
  (ICPR)}, pp.\  9227--9233. IEEE, 2021.

\bibitem[Carlucci et~al.(2019)Carlucci, D'Innocente, Bucci, Caputo, and
  Tommasi]{carlucci2019domain}
Fabio~M Carlucci, Antonio D'Innocente, Silvia Bucci, Barbara Caputo, and
  Tatiana Tommasi.
\newblock Domain generalization by solving jigsaw puzzles.
\newblock In \emph{Proceedings of the IEEE/CVF Conference on Computer Vision
  and Pattern Recognition}, pp.\  2229--2238, 2019.

\bibitem[Chang et~al.(2020)Chang, Zhang, Yu, and Jaakkola]{chang2020invariant}
Shiyu Chang, Yang Zhang, Mo~Yu, and Tommi Jaakkola.
\newblock Invariant rationalization.
\newblock In \emph{International Conference on Machine Learning}, pp.\
  1448--1458. PMLR, 2020.

\bibitem[Chen et~al.(2022)Chen, Zhou, Bian, Xie, Ma, Zhang, Yang, Han, and
  Cheng]{chen2022pareto}
Yongqiang Chen, Kaiwen Zhou, Yatao Bian, Binghui Xie, Kaili Ma, Yonggang Zhang,
  Han Yang, Bo~Han, and James Cheng.
\newblock Pareto invariant risk minimization.
\newblock \emph{arXiv preprint arXiv:2206.07766}, 2022.

\bibitem[Choe et~al.(2020)Choe, Ham, and Park]{choe2020empirical}
Yo~Joong Choe, Jiyeon Ham, and Kyubyong Park.
\newblock An empirical study of invariant risk minimization.
\newblock \emph{arXiv preprint arXiv:2004.05007}, 2020.

\bibitem[De~Haan et~al.(2019)De~Haan, Jayaraman, and Levine]{de2019causal}
Pim De~Haan, Dinesh Jayaraman, and Sergey Levine.
\newblock Causal confusion in imitation learning.
\newblock \emph{Advances in Neural Information Processing Systems}, 32, 2019.

\bibitem[DeGrave et~al.(2021)DeGrave, Janizek, and Lee]{degrave2021ai}
Alex~J DeGrave, Joseph~D Janizek, and Su-In Lee.
\newblock Ai for radiographic covid-19 detection selects shortcuts over signal.
\newblock \emph{Nature Machine Intelligence}, 3\penalty0 (7):\penalty0
  610--619, 2021.

\bibitem[Dou et~al.(2019)Dou, Coelho~de Castro, Kamnitsas, and
  Glocker]{dou2019domain}
Qi~Dou, Daniel Coelho~de Castro, Konstantinos Kamnitsas, and Ben Glocker.
\newblock Domain generalization via model-agnostic learning of semantic
  features.
\newblock \emph{Advances in Neural Information Processing Systems}, 32, 2019.

\bibitem[Dranker et~al.(2021)Dranker, He, and Belinkov]{dranker2021irm}
Yana Dranker, He~He, and Yonatan Belinkov.
\newblock Irm—when it works and when it doesn't: A test case of natural
  language inference.
\newblock \emph{Advances in Neural Information Processing Systems},
  34:\penalty0 18212--18224, 2021.

\bibitem[Foret et~al.(2020)Foret, Kleiner, Mobahi, and
  Neyshabur]{foret2020sharpness}
Pierre Foret, Ariel Kleiner, Hossein Mobahi, and Behnam Neyshabur.
\newblock Sharpness-aware minimization for efficiently improving
  generalization.
\newblock \emph{arXiv preprint arXiv:2010.01412}, 2020.

\bibitem[Ganin et~al.(2016)Ganin, Ustinova, Ajakan, Germain, Larochelle,
  Laviolette, Marchand, and Lempitsky]{ganin2016domain}
Yaroslav Ganin, Evgeniya Ustinova, Hana Ajakan, Pascal Germain, Hugo
  Larochelle, Fran{\c{c}}ois Laviolette, Mario Marchand, and Victor Lempitsky.
\newblock Domain-adversarial training of neural networks.
\newblock \emph{The journal of machine learning research}, 17\penalty0
  (1):\penalty0 2096--2030, 2016.

\bibitem[Geirhos et~al.(2020)Geirhos, Jacobsen, Michaelis, Zemel, Brendel,
  Bethge, and Wichmann]{geirhos2020shortcut}
Robert Geirhos, J{\"o}rn-Henrik Jacobsen, Claudio Michaelis, Richard Zemel,
  Wieland Brendel, Matthias Bethge, and Felix~A Wichmann.
\newblock Shortcut learning in deep neural networks.
\newblock \emph{Nature Machine Intelligence}, 2\penalty0 (11):\penalty0
  665--673, 2020.

\bibitem[Gould et~al.(2016)Gould, Fernando, Cherian, Anderson, Cruz, and
  Guo]{gould2016differentiating}
Stephen Gould, Basura Fernando, Anoop Cherian, Peter Anderson, Rodrigo~Santa
  Cruz, and Edison Guo.
\newblock On differentiating parameterized argmin and argmax problems with
  application to bi-level optimization.
\newblock \emph{arXiv preprint arXiv:1607.05447}, 2016.

\bibitem[Goyal et~al.(2017)Goyal, Doll{\'a}r, Girshick, Noordhuis, Wesolowski,
  Kyrola, Tulloch, Jia, and He]{goyal2017accurate}
Priya Goyal, Piotr Doll{\'a}r, Ross Girshick, Pieter Noordhuis, Lukasz
  Wesolowski, Aapo Kyrola, Andrew Tulloch, Yangqing Jia, and Kaiming He.
\newblock Accurate, large minibatch {SGD}: Training imagenet in 1 hour.
\newblock \emph{arXiv preprint arXiv:1706.02677}, 2017.

\bibitem[Gulrajani \& Lopez-Paz(2020)Gulrajani and
  Lopez-Paz]{gulrajani2020search}
Ishaan Gulrajani and David Lopez-Paz.
\newblock In search of lost domain generalization.
\newblock \emph{arXiv preprint arXiv:2007.01434}, 2020.

\bibitem[He et~al.(2016)He, Zhang, Ren, and Sun]{he2016deep}
Kaiming He, Xiangyu Zhang, Shaoqing Ren, and Jian Sun.
\newblock Deep residual learning for image recognition.
\newblock In \emph{Proceedings of the IEEE conference on computer vision and
  pattern recognition}, pp.\  770--778, 2016.

\bibitem[Jean et~al.(2016)Jean, Burke, Xie, Davis, Lobell, and
  Ermon]{jean2016combining}
Neal Jean, Marshall Burke, Michael Xie, W~Matthew Davis, David~B Lobell, and
  Stefano Ermon.
\newblock Combining satellite imagery and machine learning to predict poverty.
\newblock \emph{Science}, 353\penalty0 (6301):\penalty0 790--794, 2016.

\bibitem[Jin et~al.(2020)Jin, Barzilay, and Jaakkola]{jin2020domain}
Wengong Jin, Regina Barzilay, and Tommi Jaakkola.
\newblock Domain extrapolation via regret minimization.
\newblock \emph{arXiv preprint arXiv:2006.03908}, 2020.

\bibitem[Kamath et~al.(2021)Kamath, Tangella, Sutherland, and
  Srebro]{kamath2021does}
Pritish Kamath, Akilesh Tangella, Danica Sutherland, and Nathan Srebro.
\newblock Does invariant risk minimization capture invariance?
\newblock In \emph{International Conference on Artificial Intelligence and
  Statistics}, pp.\  4069--4077. PMLR, 2021.

\bibitem[Keskar et~al.(2016)Keskar, Mudigere, Nocedal, Smelyanskiy, and
  Tang]{keskar2016large}
Nitish~Shirish Keskar, Dheevatsa Mudigere, Jorge Nocedal, Mikhail Smelyanskiy,
  and Ping Tak~Peter Tang.
\newblock On large-batch training for deep learning: Generalization gap and
  sharp minima.
\newblock \emph{arXiv preprint arXiv:1609.04836}, 2016.

\bibitem[Kingma \& Ba(2014)Kingma and Ba]{kingma2014adam}
Diederik~P Kingma and Jimmy Ba.
\newblock Adam: A method for stochastic optimization.
\newblock \emph{arXiv preprint arXiv:1412.6980}, 2014.

\bibitem[Koh et~al.(2021)Koh, Sagawa, Marklund, Xie, Zhang, Balsubramani, Hu,
  Yasunaga, Phillips, Gao, et~al.]{koh2021wilds}
Pang~Wei Koh, Shiori Sagawa, Henrik Marklund, Sang~Michael Xie, Marvin Zhang,
  Akshay Balsubramani, Weihua Hu, Michihiro Yasunaga, Richard~Lanas Phillips,
  Irena Gao, et~al.
\newblock Wilds: A benchmark of in-the-wild distribution shifts.
\newblock In \emph{International Conference on Machine Learning}, pp.\
  5637--5664. PMLR, 2021.

\bibitem[Krizhevsky et~al.(2017)Krizhevsky, Sutskever, and
  Hinton]{krizhevsky2017imagenet}
Alex Krizhevsky, Ilya Sutskever, and Geoffrey~E Hinton.
\newblock Imagenet classification with deep convolutional neural networks.
\newblock \emph{Communications of the ACM}, 60\penalty0 (6):\penalty0 84--90,
  2017.

\bibitem[Krueger et~al.(2021)Krueger, Caballero, Jacobsen, Zhang, Binas, Zhang,
  Le~Priol, and Courville]{krueger2021out}
David Krueger, Ethan Caballero, Joern-Henrik Jacobsen, Amy Zhang, Jonathan
  Binas, Dinghuai Zhang, Remi Le~Priol, and Aaron Courville.
\newblock Out-of-distribution generalization via risk extrapolation (rex).
\newblock In \emph{International Conference on Machine Learning}, pp.\
  5815--5826. PMLR, 2021.

\bibitem[Li et~al.(2017)Li, Yang, Song, and Hospedales]{li2017deeper}
Da~Li, Yongxin Yang, Yi-Zhe Song, and Timothy~M Hospedales.
\newblock Deeper, broader and artier domain generalization.
\newblock In \emph{Proceedings of the IEEE international conference on computer
  vision}, pp.\  5542--5550, 2017.

\bibitem[Li et~al.(2019)Li, Zhang, Yang, Liu, Song, and
  Hospedales]{li2019episodic}
Da~Li, Jianshu Zhang, Yongxin Yang, Cong Liu, Yi-Zhe Song, and Timothy~M
  Hospedales.
\newblock Episodic training for domain generalization.
\newblock In \emph{Proceedings of the IEEE/CVF International Conference on
  Computer Vision}, pp.\  1446--1455, 2019.

\bibitem[Li et~al.(2018)Li, Xu, Taylor, Studer, and
  Goldstein]{li2018visualizing}
Hao Li, Zheng Xu, Gavin Taylor, Christoph Studer, and Tom Goldstein.
\newblock Visualizing the loss landscape of neural nets.
\newblock \emph{Advances in neural information processing systems}, 31, 2018.

\bibitem[Lin et~al.(2021)Lin, Lian, and Zhang]{lin2021empirical}
Yong Lin, Qing Lian, and Tong Zhang.
\newblock An empirical study of invariant risk minimization on deep models.
\newblock In \emph{ICML 2021 Workshop on Uncertainty and Robustness in Deep
  Learning}, pp.\ ~7, 2021.

\bibitem[Lin et~al.(2022)Lin, Dong, Wang, and Zhang]{lin2022bayesian}
Yong Lin, Hanze Dong, Hao Wang, and Tong Zhang.
\newblock Bayesian invariant risk minimization.
\newblock In \emph{Proceedings of the IEEE/CVF Conference on Computer Vision
  and Pattern Recognition}, pp.\  16021--16030, 2022.

\bibitem[Liu et~al.(2021)Liu, Gao, Zhang, Meng, and Lin]{liu2021investigating}
Risheng Liu, Jiaxin Gao, Jin Zhang, Deyu Meng, and Zhouchen Lin.
\newblock Investigating bi-level optimization for learning and vision from a
  unified perspective: A survey and beyond.
\newblock \emph{IEEE Transactions on Pattern Analysis and Machine
  Intelligence}, 2021.

\bibitem[Liu et~al.(2015)Liu, Luo, Wang, and Tang]{liu2015faceattributes}
Ziwei Liu, Ping Luo, Xiaogang Wang, and Xiaoou Tang.
\newblock Deep learning face attributes in the wild.
\newblock In \emph{Proceedings of International Conference on Computer Vision
  (ICCV)}, December 2015.

\bibitem[Long et~al.(2015)Long, Cao, Wang, and Jordan]{long2015learning}
Mingsheng Long, Yue Cao, Jianmin Wang, and Michael Jordan.
\newblock Learning transferable features with deep adaptation networks.
\newblock In \emph{International conference on machine learning}, pp.\
  97--105. PMLR, 2015.

\bibitem[Nam et~al.(2021)Nam, Lee, Park, Yoon, and Yoo]{nam2021reducing}
Hyeonseob Nam, HyunJae Lee, Jongchan Park, Wonjun Yoon, and Donggeun Yoo.
\newblock Reducing domain gap by reducing style bias.
\newblock In \emph{Proceedings of the IEEE/CVF Conference on Computer Vision
  and Pattern Recognition}, pp.\  8690--8699, 2021.

\bibitem[Peters et~al.(2016)Peters, B{\"u}hlmann, and
  Meinshausen]{peters2016causal}
Jonas Peters, Peter B{\"u}hlmann, and Nicolai Meinshausen.
\newblock Causal inference by using invariant prediction: identification and
  confidence intervals.
\newblock \emph{Journal of the Royal Statistical Society: Series B (Statistical
  Methodology)}, 78\penalty0 (5):\penalty0 947--1012, 2016.

\bibitem[Rame et~al.(2022)Rame, Dancette, and Cord]{rame2022fishr}
Alexandre Rame, Corentin Dancette, and Matthieu Cord.
\newblock Fishr: Invariant gradient variances for out-of-distribution
  generalization.
\newblock In \emph{International Conference on Machine Learning}, pp.\
  18347--18377. PMLR, 2022.

\bibitem[Rosenfeld et~al.(2020)Rosenfeld, Ravikumar, and
  Risteski]{rosenfeld2020risks}
Elan Rosenfeld, Pradeep Ravikumar, and Andrej Risteski.
\newblock The risks of invariant risk minimization.
\newblock \emph{arXiv preprint arXiv:2010.05761}, 2020.

\bibitem[Sagawa et~al.(2019)Sagawa, Koh, Hashimoto, and
  Liang]{sagawa2019distributionally}
Shiori Sagawa, Pang~Wei Koh, Tatsunori~B Hashimoto, and Percy Liang.
\newblock Distributionally robust neural networks for group shifts: On the
  importance of regularization for worst-case generalization.
\newblock \emph{arXiv preprint arXiv:1911.08731}, 2019.

\bibitem[Sagawa et~al.(2020)Sagawa, Raghunathan, Koh, and
  Liang]{sagawa2020investigation}
Shiori Sagawa, Aditi Raghunathan, Pang~Wei Koh, and Percy Liang.
\newblock An investigation of why overparameterization exacerbates spurious
  correlations.
\newblock In \emph{International Conference on Machine Learning}, pp.\
  8346--8356. PMLR, 2020.

\bibitem[Shah et~al.(2020)Shah, Tamuly, Raghunathan, Jain, and
  Netrapalli]{shah2020pitfalls}
Harshay Shah, Kaustav Tamuly, Aditi Raghunathan, Prateek Jain, and Praneeth
  Netrapalli.
\newblock The pitfalls of simplicity bias in neural networks.
\newblock \emph{Advances in Neural Information Processing Systems},
  33:\penalty0 9573--9585, 2020.

\bibitem[Simonyan \& Zisserman(2014)Simonyan and Zisserman]{simonyan2014very}
Karen Simonyan and Andrew Zisserman.
\newblock Very deep convolutional networks for large-scale image recognition.
\newblock \emph{arXiv preprint arXiv:1409.1556}, 2014.

\bibitem[Sun et~al.(2014)Sun, Wang, and Tang]{sun2014deep}
Yi~Sun, Xiaogang Wang, and Xiaoou Tang.
\newblock Deep learning face representation from predicting 10,000 classes.
\newblock In \emph{Proceedings of the IEEE conference on computer vision and
  pattern recognition}, pp.\  1891--1898, 2014.

\bibitem[Torralba \& Efros(2011)Torralba and Efros]{torralba2011unbiased}
Antonio Torralba and Alexei~A Efros.
\newblock Unbiased look at dataset bias.
\newblock In \emph{CVPR 2011}, pp.\  1521--1528. IEEE, 2011.

\bibitem[Tzeng et~al.(2014)Tzeng, Hoffman, Zhang, Saenko, and
  Darrell]{tzeng2014deep}
Eric Tzeng, Judy Hoffman, Ning Zhang, Kate Saenko, and Trevor Darrell.
\newblock Deep domain confusion: Maximizing for domain invariance.
\newblock \emph{arXiv preprint arXiv:1412.3474}, 2014.

\bibitem[Wang et~al.(2022)Wang, Lan, Liu, Ouyang, Qin, Lu, Chen, Zeng, and
  Yu]{wang2022generalizing}
Jindong Wang, Cuiling Lan, Chang Liu, Yidong Ouyang, Tao Qin, Wang Lu, Yiqiang
  Chen, Wenjun Zeng, and Philip Yu.
\newblock Generalizing to unseen domains: A survey on domain generalization.
\newblock \emph{IEEE Transactions on Knowledge and Data Engineering}, 2022.

\bibitem[Xie et~al.(2020)Xie, Ye, Chen, Liu, Sun, and Li]{xie2020risk}
Chuanlong Xie, Haotian Ye, Fei Chen, Yue Liu, Rui Sun, and Zhenguo Li.
\newblock Risk variance penalization.
\newblock \emph{arXiv preprint arXiv:2006.07544}, 2020.

\bibitem[Xu et~al.(2022)Xu, Zhang, Cui, Li, Shen, and Xu]{xu2022regulatory}
Renzhe Xu, Xingxuan Zhang, Peng Cui, Bo~Li, Zheyan Shen, and Jiazheng Xu.
\newblock Regulatory instruments for fair personalized pricing.
\newblock In \emph{Proceedings of the ACM Web Conference 2022}, pp.\  4--15,
  2022.

\bibitem[Xu \& Jaakkola(2021)Xu and Jaakkola]{xu2021learning}
Yilun Xu and Tommi Jaakkola.
\newblock Learning representations that support robust transfer of predictors.
\newblock \emph{arXiv preprint arXiv:2110.09940}, 2021.

\bibitem[You et~al.(2017{\natexlab{a}})You, Gitman, and Ginsburg]{you2017large}
Yang You, Igor Gitman, and Boris Ginsburg.
\newblock Large batch training of convolutional networks.
\newblock \emph{arXiv preprint arXiv:1708.03888}, 2017{\natexlab{a}}.

\bibitem[You et~al.(2017{\natexlab{b}})You, Gitman, and
  Ginsburg]{you2017scaling}
Yang You, Igor Gitman, and Boris Ginsburg.
\newblock Scaling {SGD} batch size to 32k for imagenet training.
\newblock \emph{arXiv preprint arXiv:1708.03888}, 6, 2017{\natexlab{b}}.

\bibitem[You et~al.(2018)You, Zhang, Hsieh, Demmel, and
  Keutzer]{you2018imagenet}
Yang You, Zhao Zhang, Cho-Jui Hsieh, James Demmel, and Kurt Keutzer.
\newblock Imagenet training in minutes.
\newblock In \emph{Proceedings of the 47th International Conference on Parallel
  Processing}, pp.\ ~1. ACM, 2018.

\bibitem[You et~al.(2019)You, Li, Reddi, Hseu, Kumar, Bhojanapalli, Song,
  Demmel, Keutzer, and Hsieh]{you2019large}
Yang You, Jing Li, Sashank Reddi, Jonathan Hseu, Sanjiv Kumar, Srinadh
  Bhojanapalli, Xiaodan Song, James Demmel, Kurt Keutzer, and Cho-Jui Hsieh.
\newblock Large batch optimization for deep learning: Training bert in 76
  minutes.
\newblock \emph{arXiv preprint arXiv:1904.00962}, 2019.

\bibitem[Zhang et~al.(2021)Zhang, Ahuja, Xu, Wang, and Courville]{zhang2021can}
Dinghuai Zhang, Kartik Ahuja, Yilun Xu, Yisen Wang, and Aaron Courville.
\newblock Can subnetwork structure be the key to out-of-distribution
  generalization?
\newblock In \emph{International Conference on Machine Learning}, pp.\
  12356--12367. PMLR, 2021.

\bibitem[Zhang et~al.(2022{\natexlab{a}})Zhang, Lu, Zhang, Chen, Chen, Fan,
  Martie, Horesh, Hong, and Liu]{zhang2022distributed}
Gaoyuan Zhang, Songtao Lu, Yihua Zhang, Xiangyi Chen, Pin-Yu Chen, Quanfu Fan,
  Lee Martie, Lior Horesh, Mingyi Hong, and Sijia Liu.
\newblock Distributed adversarial training to robustify deep neural networks at
  scale.
\newblock In \emph{Uncertainty in Artificial Intelligence}, pp.\  2353--2363.
  PMLR, 2022{\natexlab{a}}.

\bibitem[Zhang et~al.(2022{\natexlab{b}})Zhang, Zhou, Xu, Cui, Shen, and
  Liu]{zhang2022nico}
Xingxuan Zhang, Linjun Zhou, Renzhe Xu, Peng Cui, Zheyan Shen, and Haoxin Liu.
\newblock Nico++: Towards better benchmarking for domain generalization.
\newblock \emph{arXiv preprint arXiv:2204.08040}, 2022{\natexlab{b}}.

\bibitem[Zhou et~al.(2022{\natexlab{a}})Zhou, Liu, Qiao, Xiang, and
  Loy]{zhou2022domain}
Kaiyang Zhou, Ziwei Liu, Yu~Qiao, Tao Xiang, and Chen~Change Loy.
\newblock Domain generalization: A survey.
\newblock \emph{IEEE Transactions on Pattern Analysis and Machine
  Intelligence}, 2022{\natexlab{a}}.

\bibitem[Zhou et~al.(2022{\natexlab{b}})Zhou, Lin, Zhang, and
  Zhang]{zhou2022sparse}
Xiao Zhou, Yong Lin, Weizhong Zhang, and Tong Zhang.
\newblock Sparse invariant risk minimization.
\newblock In \emph{International Conference on Machine Learning}, pp.\
  27222--27244. PMLR, 2022{\natexlab{b}}.

\end{thebibliography}
